\documentclass[11pt]{article}

\usepackage[preprint]{colm2026_conference}

\usepackage{lineno}
\usepackage{hyperref}
\usepackage{url}

\usepackage[T1]{fontenc}
\usepackage[utf8]{inputenc}

\usepackage{microtype}
\usepackage{inconsolata}

\usepackage{booktabs}
\usepackage{tabularx}
\usepackage{array}
\usepackage{longtable}
\usepackage{graphicx}
\usepackage{subcaption}

\usepackage{amsfonts}
\usepackage{amsmath}

\usepackage{tikz}
\usetikzlibrary{arrows.meta,positioning,shapes.multipart,fit}

\usepackage{enumitem}
\usepackage{xspace}
\usepackage{mlpaper_visual_style}

\newcommand{\numpapers}{223\xspace}

\newcommand{\reffig}[1]{Figure~\ref{fig:#1}}
\newcommand{\reftab}[1]{Table~\ref{tab:#1}}
\newcommand{\refsec}[1]{Section~\ref{sec:#1}}
\newcommand{\refapp}[1]{Appendix~\ref{sec:#1}}

\newif\ifsubmission
\submissionfalse

\newcolumntype{Y}{>{\raggedright\arraybackslash}X}

\title{Shall We Play a Game?\\ Language Models for Open-ended Wargames}

\usepackage{twemojis}
\newcommand{\AffGT}{\twemoji{bee}}          
\newcommand{\AffGTSafety}{\twemoji{shield}} 
\newcommand{\AffMATS}{\twemoji{fire}}       

\ifsubmission
\author{%
  Anonymous Authors
}
\else
\author{%
  Glenn Matlin$^{\AffGT\,\AffGTSafety\,\AffMATS}$ \\
  Isaac Song$^{\AffGT\,*}$ \quad Yixiong Hao$^{\AffGT\,\AffGTSafety\,*}$ \quad Parv Mahajan$^{\AffGT\,\AffGTSafety\,*}$ \\
  Evan Montoya$^{\AffGT\,\dagger}$ \quad Ryan Bard$^{\AffGT\,\dagger}$ \quad Stuart R. Topp$^{\AffGT\,\dagger}$ \\
  Anthony Wen-Ming Zang$^{\AffGT\,\ddagger}$ \quad Mohammed Rehan Parwani$^{\AffGT\,\ddagger}$ \quad Soham Shetty$^{\AffGT\,\ddagger}$ \\
  Mark Riedl$^{\AffGT}$ \\[4pt]
  \AffGT\,College of Computing, Georgia Institute of Technology \\
  \AffGTSafety\,Georgia Tech AI Safety Initiative \quad
  \AffMATS\,MATS Program
}
\fi

\begin{document}

\ifcolmsubmission
  \linenumbers
\fi

\maketitle

\ifsubmission\else
  \lhead{Published at the Social Sim'26 Workshop, COLM 2026}
\fi

\begin{abstract}
LLM-based social simulations can make a generated transcript look like a single behavioral signal, but the model behind that transcript may be doing several different jobs: choosing what an actor says or does, deciding what happens after an action, or both. The difference matters especially in open-ended wargames, where models are prized for handling unusual actions and ambiguous consequences. We report a scoping review of \numpapers~de-duplicated AI-in-wargames and strategic-simulation papers retrieved through May 1, 2026, describing each simulation by its \emph{model-control profile}: whether the language model has open-ended control over player actions, adjudication, or both. Only 20 of \numpapers~studies ($\approx$9\%) give language models both roles. Before treating LM outputs as social simulations, researchers need to know how much creative control the model has over actions and consequences. For open-ended simulations, fidelity depends not only on whether agents behave plausibly, but also on whether language models can reliably act as adjudicators or world models.
\end{abstract}

\section{Introduction}
\label{sec:introduction}
A social simulation can look plausible for the wrong reason. A transcript may show agents negotiating, escalating, compromising, or planning, but the reader still needs to know where that behavior came from. Did the model choose the participant's action, or did a rule system constrain the result? And who decided what the action caused? Without that separation, the transcript is hard to interpret as evidence about social behavior.

Wargames make this problem unusually clear. A serious wargame simulates conflict under uncertainty to elicit judgment about plans, assumptions, adversaries, and possible futures \citep{perla_what_wargaming_1985, rubel_epistemology_war_2006, wallman_its_only_1995, us_army_war_college_strategic_wargaming_2015}. Military examples are the familiar ones, but organizations run wargaming-like simulations for diplomacy, business, public health, cybersecurity, law, and other settings where participant decisions shape later outcomes. Language models are attractive here: they can play roles, generate scenario branches, adjudicate ambiguous actions, and summarize lessons \citep{hogan_openended_wargames_2024}. That same flexibility makes them difficult to evaluate, because once language can influence both what an actor attempts and what later becomes true in the simulated world, a fluent storyline is no longer enough.

The key issue is not whether a model is ``open-ended'' in general but where the openness sits. A model that proposes a surprising move inside a rule-bound game creates a different kind of evidence from a model that also decides the consequences of that move. In the second setup, language holds authority over the simulated world itself, which is useful for studying improvisation, escalation, and negotiation but raises sharper validity and safety questions. Prior work already reports brittle reasoning, hallucinations, rule violations, and escalation risks in LM games and diplomatic simulations \citep{gao_large_language_2024, ma_computational_experiments_2024, feng_survey_large_2025, yao_spinbench_how_2025, lamparth_human_vs_2024, rivera_escalation_risks_2024}. A social-simulation evaluation should therefore ask not only what agents say, but which role gives language control over consequences.

We call this a simulation's \emph{model-control profile}: whether language models have open-ended control over player actions, adjudication, or both. This profile operationalizes \emph{role-specific open-endedness}, the distinction our rubric measures.

As simulations move from constrained benchmarks toward open-ended settings, where a model can turn an utterance into a new world state, the stakes rise. The safety concern is not only that a model might say something implausible, but that the simulation may accept unsupported language as a consequence and build later social behavior on top of it.

We use LM wargames as a concrete testbed for this broader question: what does the model control? We make three contributions. First, we curate a catalog of \numpapers de-duplicated AI-in-wargames and wargame-relevant strategic-simulation studies retrieved through May 1, 2026. Second, we introduce a two-axis rubric that separates open-ended player action from open-ended adjudication, with decision rules and worked examples for each quadrant. Third, we use the rubric to map the literature and show that the both-creative regime remains rare: only 20 of \numpapers~studies place language models in both player and adjudicator roles. This is a scoping review by design: it maps where model control sits, and makes no claims about model performance in either role.

\section{Background}
\label{sec:background}
\begin{figure*}[t]
\centering
\begin{subfigure}[t]{0.40\linewidth}
\centering
\includegraphics[width=\linewidth]{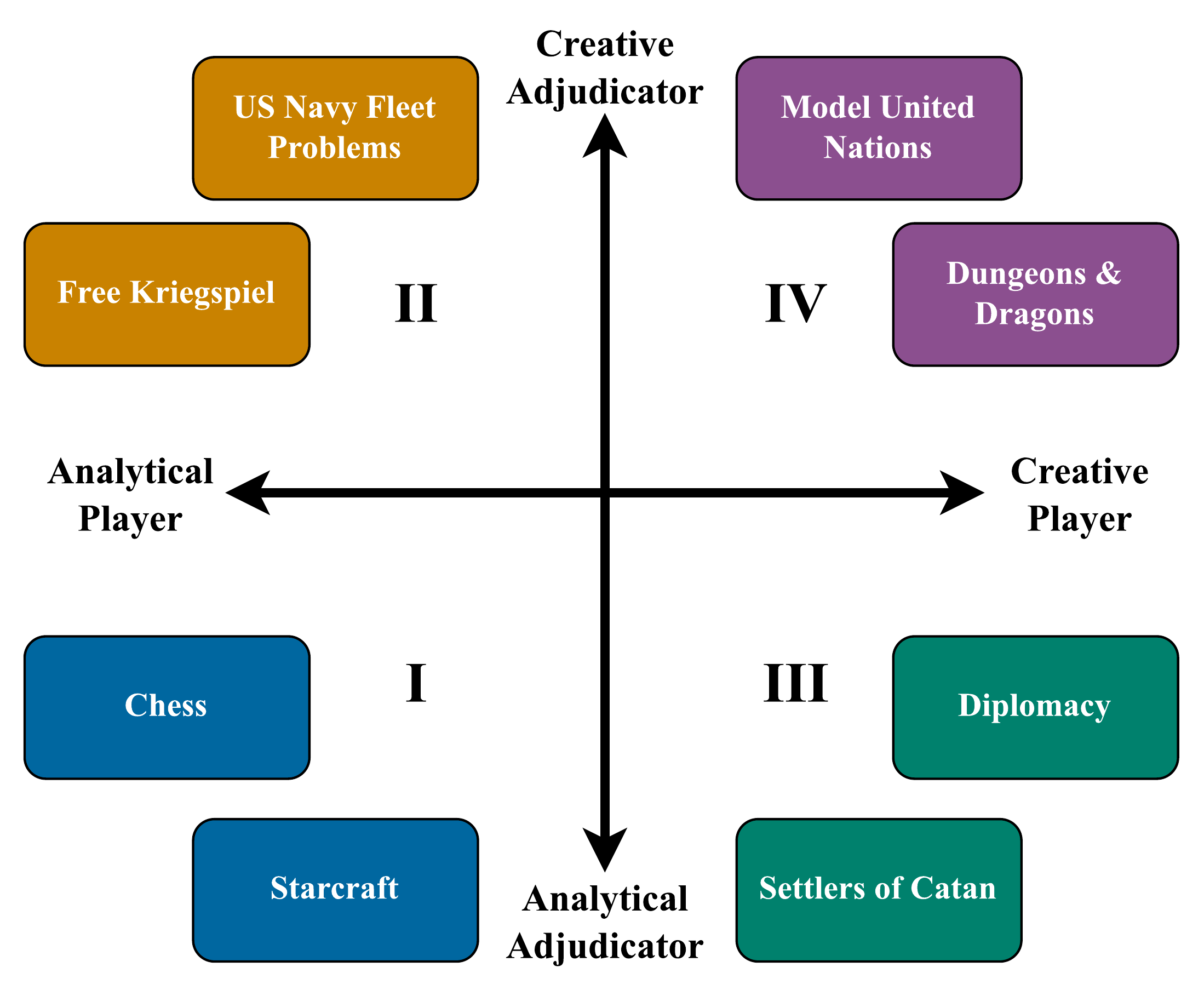}
\caption{Two-axis rubric with illustrative example games per quadrant.}
\label{fig:creativity-quadrants}
\end{subfigure}\hfill
\begin{subfigure}[t]{0.58\linewidth}
\centering
\includegraphics[width=\linewidth]{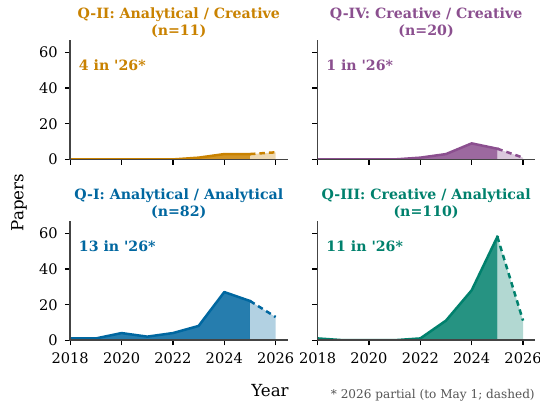}
\caption{Per-quadrant main-corpus paper counts over time (2018--2026), with panels positioned as in the rubric. The dashed, lighter 2026 segment is partial (papers through May 1, 2026), not a decline. Q-III rises sharply while Q-IV (both-creative) stays rare.}
\label{fig:plot-over-time}
\end{subfigure}
\caption{Role-specific open-endedness: (a) the rubric and (b) the full \numpapers-paper main corpus by quadrant over time. The corpus was retrieved through May 1, 2026, so the 2026 counts in panel (b) are partial.}
\label{fig:overview}
\end{figure*}

We focus on open-ended serious wargames \citep{perla_what_wargaming_1985, coulthart_whats_problem_2017, rubel_epistemology_war_2006, morgan_wargames_training_1991}. A serious game is designed for more than entertainment; its aims include training, education, analysis, or policy exploration \citep{wallman_its_only_1995, smith_serious_games_2020}, and unlike recreational games its design is built to generate insight and support learning \citep{de_rosa_design_methodology_2021}. Serious wargames bring this stance to conflict: experts use them to explore strategic choices, gather judgment, or stress-test plans against a simulated adversary \citep{us_army_war_college_strategic_wargaming_2015}.

\subsection{Open-Endedness}
\label{sec:open-endedness}

Open-endedness has been argued to be essential for AI systems that must demonstrate genuine novelty, adaptation, and continual learning rather than excel only on fixed benchmarks \citep{hughes_openendedness_essential_2024, stanley_why_openendedness_2019}; serious wargaming sits squarely in this regime. The idea has a long lineage, running from evolutionary computation and artificial life through to AI systems that generate unpredictable outputs, behaviors, and artifacts without predetermined limits \citep{sigaud_definition_openended_2024, stepney_openendedness_detecting_2024, taylor_openended_evolution_2016, packard_overview_openended_2019, soros_identifying_necessary_2014, stanley_why_greatness_2015, soros_openendedness_last_2017}. For this review, we define open-endedness as a system's capacity to keep producing novel, evolving outcomes without settling into predictable patterns or repeating cycles.

Wargames are a natural setting for open-endedness, since participants generate novel actions, strategies, and narratives over long interactions \citep{samvelyan_maestro_openended_2023, zhang_omni_openendedness_2024, faldor_omniepic_openendedness_2025}. The open-endedness is asymmetric and role-dependent, though. Players may innovate freely while adjudicators vary in outcome flexibility (see \refsec{player_vs_adjudicator} and \refsec{creative_vs_analytical}), so it can surface as player creativity, adjudicator creativity, or both. Our rubric (\reffig{creativity-quadrants}) makes these dimensions explicit and links linguistic agents to wider strategy and outcome spaces.

\begin{wgrubricbox}{Protocol-level coding rule}
We code what the game protocol permits, not only what happened in one observed playthrough. Natural-language commentary around fixed moves remains \lowcode; a terse move remains \highcode when the protocol lets the actor introduce unenumerated proposals or consequences.
\end{wgrubricbox}

Within this review, we treat AI involvement in wargames as open-ended when the move space is not fully predetermined, scenarios accommodate dynamic goals such as shifting alliances or objectives, adversaries or stakeholders adapt over time, adjudication can address unforeseen actions and consequences, or the scenario set can expand through structured generation.

\subsection{Players vs Adjudicators}
\label{sec:player_vs_adjudicator}

Agents in wargames are commonly grouped into one of two categories: \emph{player} and \emph{adjudicator}. A \textbf{player} is an agent that engages in analysis and planning, proposes moves during play, and experiences the consequences of those outcomes. The \textbf{adjudicator} is an agent that determines outcomes (via rules, computation, or expert judgment) of players' moves. In some games, such as \gameChess or \gameCivilization, the players are constrained to pre-defined sets of rules. In other games, such as \gameDnD or \gameModelUN, actions are unconstrained, and a player can propose any move that they can express in natural language \citep{martin_dungeons_dqns_2018}. Likewise, while adjudicators may be constrained to predefined rules and calculations (e.g., game mechanics defining valid actions and outcomes), they might also be able to propose any effect or world change that can be expressed in natural languages (e.g., expressing player outcomes using narrative generation or mental heuristics) \citep{zhu_calypso_llms_2023}. In some cases, players and adjudicators can negotiate the effects of moves, with players making the case for why an effect could or could not occur \citep{callison-burch_dungeons_dragons_2022, zhu_fireball_dataset_2023}.

\subsection{Creative vs Analytical Wargames}
\label{sec:creative_vs_analytical}

Game design inherently trades off realism and simplicity to manage complexity \citep{wallman_its_only_1995, wasser_gaming_gray_2019, reddie_wargames_data_2023}. Therefore, game designers make intentional choices about the game's scope and rules to focus on the key skills they want to reinforce or gather data about \citep{rouse_iii_game_design_2004, walton_developing_theory_2019, booth_wargame_2024}. SMEs often classify wargames along a spectrum from \textbf{creative wargames} to \textbf{analytical wargames} \citep{zegers_matrix_game_2011, franken_when_it_2013, barzashka_five_theoretical_2019, combe_ii_educational_wargaming_2021, de_rosa_design_methodology_2021}.\footnote{In the military domain, these game formats are commonly referred to as seminar wargames and matrix wargames, respectively \citep{us_naval_war_college_war_gamers_2015, ashdown_matrix_games_2018}.} Creative games are open-ended, non-linear, and rely heavily on creative problem solving, while analytical games are heavily structured and rule-based \citep{zegers_matrix_game_2011}.\footnote{Note that the use of the term \emph{agent} creativity is not consistent with literature on \emph{human} creativity, as even playing a highly constrained game like \gameChess can be regarded as requiring creative problem solving \citep{colton_simon_computational_creativity_2012}.}

Contemporary AI research has historically focused on analytical games, in which world states and transitions are defined by strict rules (e.g., \gameChess, \gameGo, \gameStarCraft) and solved via reinforcement learning or state-space search \citep{silver_mastering_game_2017, vinyals_grandmaster_level_2019, goodman_ai_wargaming_2020, ma_adaptive_command_2025}. While these AI techniques are powerful solvers \citep{wan_alphazerolike_treesearch_2023, black_scaling_artificial_2024}, they often sacrifice generalization to novel scenarios \citep{balloch_novgrid_flexible_2022} and do not present opportunities to incorporate the ambiguity, creativity, or multi-party interactions central to open-ended games. \refsec{methodology} operationalizes this design space with separate player- and adjudicator-creativity axes.

\section{Related Work}
\label{sec:related-work}
LLM-based social simulation is developing across several adjacent threads. Social-science uses of ``silicon samples'' ask whether conditioned language models can reproduce human response distributions, framing the key validity question as one of algorithmic fidelity \citep{argyle_out_one_2022}. Generative-agent systems now range from small interactive sandboxes to large many-agent worlds and reusable simulation libraries \citep{park_generative_agents_2023, al_project_sid_2024, vezhnevets_generative_agentbased_2023}. Concordia is especially relevant here because agents describe actions in natural language while a game-master-like mediator decides what happens next; later guidance using Concordia argues that validation has to be built into the experimental protocol, not checked only after a run \citep{vezhnevets_generative_agentbased_2023, navarro_designing_reliable_2024}. Recent work shows why that caution matters: LLM populations can develop conventions and collective biases, while LMs used as human replacements can flatten identity groups \citep{ashery_emergent_social_2025, wang_large_language_2025}. Surveys of LLM-empowered agent-based modeling likewise identify action generation, environment perception, human alignment, and evaluation as central challenges for simulation \citep{gao_large_language_2024}. Interactive benchmarks such as SOTOPIA evaluate social intelligence in open-ended role-play settings \citep{zhou2023sotopia}, while work on text-based world simulators directly tests whether language models can predict how actions change world states \citep{Wang2024serve}. Our review connects these concerns in a setting where social action and world-state updates may be assigned to different model roles. The model-control profile of a simulation is therefore part of the validity question, not an implementation detail.

Our work is also related to surveys of LMs in games: agent architecture \citep{hu_survey_large_2025}, modeling \citep{gao_large_language_2024}, simulation \citep{ma_computational_experiments_2024}, and general game applications \citep{xu_survey_game_2024}. There are also survey papers on strategic reasoning \citep{zhang_llm_mastermind_2024}, game theory \citep{sun_game_theory_2025}, and benchmarking social reasoning and persuasion \citep{yao_spinbench_how_2025, park_ai_deception_2024, feng_survey_large_2025}.

These surveys often focus on games with well-defined rules and clear win conditions. In contrast, our work focuses on the sub-domain of open-ended, language-based wargames characterized by their qualitative nature, ambiguous rules, and the importance of narrative and argumentation. To our knowledge, this is the first archival scoping literature review of LM-era wargaming together with an operational coding rubric for adjudicator- and player-side open-endedness.

\section{Methodology}
\label{sec:methodology}
We conducted a scoping literature review \citep{arksey_scoping_studies_2005} of unclassified work on AI in wargames and wargame-relevant strategic simulations. The purpose of the review was to map the literature's coverage rather than estimate effect sizes: we aimed to identify how AI systems are used as players, adjudicators, world models, state generators, facilitators, analysts, or other decision-support components in interactive strategic environments, and to code those studies using our role-specific open-endedness rubric. The review follows the logic of scoping studies, using broad retrieval, iterative screening, and transparent categorization to map a heterogeneous research area.

\paragraph{Search strategy and candidate retrieval.}
Candidate retrieval combined broad database coverage with high precision through a hybrid workflow. A custom Python program harvested academic databases, translating query families, normalizing records, and fetching preprints; every surviving record then went through manual validation and coding. We ran the searches between \textbf{November 1, 2025 and May 1, 2026} across Google Scholar, Semantic Scholar, and arXiv, with no hard publication-date filter at retrieval. Any study matching the eligibility criteria below entered the catalog regardless of release year, which let earlier methodological, definitional, and historical work surface alongside contemporary records. The full main corpus contains \numpapers~de-duplicated studies retrieved through the May 1, 2026 cutoff.

Searches combined four concept groups (wargaming environments, AI/LM methods, AI roles, and strategic interaction terms). Since indexing services differ in their support for Boolean syntax, we wrote query families rather than a single query and let the program expand each family into API-optimized variants. We supplemented automated retrieval with backward and forward snowballing from seed papers and wargaming handbooks, and tracked every candidate record's eligibility and coding in a spreadsheet. \refapp{search} reports the verbatim query families, program phases, and catalog contents.

\paragraph{Eligibility screening.}
Candidate papers were first screened by title and abstract, then reviewed in full text when the title or abstract was ambiguous or potentially relevant. We anchored screening on a definition of ``Artificial Intelligence'' (AI) old enough to predate any particular method family: ``the use of computers to carry out tasks that previously required human intelligence'' \citep{minsky_semantic_1968}. The breadth is deliberate. LMs drive much of the recent excitement in AI for wargaming, but the survey is not restricted to them.

Four criteria governed eligibility. The full text had to be publicly available online. The paper had to study an AI system in a broad sense: language models, reinforcement learning (single- or multi-agent), symbolic AI, expert systems, hybrid systems, or other machine-learning approaches. It had to involve an interactive game, wargame, simulation, exercise, or strategic environment in which agent decisions affected later state, outcomes, rewards, narratives, beliefs, or other agents. And the AI had to occupy at least one game-relevant role, such as player, adjudicator, world model, state generator, facilitator, analyst, teammate, opponent, or strategic decision module.

Exclusions mirror these criteria: records lacking full text, involving no AI, involving no interactive game or simulation, using only single-turn polling or static scenario classification without stateful consequences, duplicating another record, or amounting to editorials, blog posts, or opinion pieces without research methodology. Some excluded and near-miss records stayed in the spreadsheet anyway, as an audit trail and as background where they helped with definitions, safety discussion, or methodological framing.

\paragraph{De-duplication and canonicalization.}
We de-duplicated records on DOI, arXiv identifier, OpenReview identifier, and normalized title. A conference version, arXiv version, workshop version, and/or preprint describing the same study count as a single canonical record unless a later version added a materially distinct experiment; where a title changed between preprint and publication, we kept the most stable or published title and noted the alternate. The final set of \numpapers de-duplicated main-corpus studies appears in \refapp{wargames}.

\paragraph{Manual post-hoc stratification.}
After eligibility screening, we manually flagged how each reviewed record was used in the paper, stratifying the retrieved records into core wargame studies, wargame-relevant proxy studies, background studies, and excluded records (\refapp{search} gives the definitions).

Only the core wargame and wargame-relevant proxy papers that satisfied the main-corpus criteria enter the quadrant analysis. Human authors made every final inclusion, exclusion, stratification, and quadrant decision; language models assisted only with metadata organization and candidate triage, and their outputs were never authoritative.

\subsection{Operational Rubric}
\label{sec:rubric}

We code each paper in our corpus along the two axes of role-specific open-endedness defined in \reftab{rubric}: \emph{player creativity} (LOW/HIGH) and \emph{adjudicator creativity} (LOW/HIGH). What matters for a code is what the wargame's protocol \emph{permits}. ``Creativity'' here is shorthand for that protocol-level openness---how large an action space the player may draw on, and how large a consequence space the adjudicator may rule over---not a judgment of creative quality. A player whose protocol admits free-form natural language is coded HIGH even when observed inputs are short, and the two axes move independently, so a paper coded HIGH on player creativity (e.g., free-form negotiation) can still be LOW on adjudicator creativity when deterministic rules settle the outcomes; \gameNoPressDiplomacy is the canonical case. The released artifact spreadsheet additionally records application domain via per-paper flags (e.g., social simulation, social debate, market/finance, military wargame), though we do not foreground domain coverage in this paper's analysis. \refapp{facets} reports three further \emph{indicative}, AI-assisted facets (AI role, method family, and evaluation approach, cross-validated across two model families) that inform the discussion but sit outside the rubric.

One author labeled each paper with LM assistance, and a second author then independently reviewed every quadrant assignment manually without LM assistance; disagreements were resolved by examining abstracts and matching against existing taxonomy anchors. We did not compute a formal inter-coder reliability statistic (e.g., Cohen's $\kappa$), since the second pass was a review of the first author's labels rather than fully independent re-coding; double-coding a stratified subset to report per-axis reliability is left to future work.

\begin{table*}[t]
\caption{Operational coding rubric for role-specific open-endedness. Coders assign each axis from the game protocol: what the protocol permits players or adjudicators to do, not what happened in one observed playthrough.}
\label{tab:rubric}
\centering
\footnotesize
\setlength{\tabcolsep}{4.5pt}
\renewcommand{\arraystretch}{1.12}
\begin{tabularx}{\textwidth}{@{}>{\raggedright\arraybackslash}p{0.15\textwidth}Y Y >{\raggedright\arraybackslash}p{0.20\textwidth}@{}}
\toprule
\textbf{Axis} & \textbf{Code \lowcode when\ldots} & \textbf{Code \highcode when\ldots} & \textbf{Boundary cue} \\
\midrule
\textbf{Player creativity}
&
\cellcolor{WargameQIBg} The player acts through a fixed action set, menu, command grammar, or structured orders. Examples include move interfaces in \gameChess and tactical military orders in \gameFreeKriegsspielQII.
&
\cellcolor{WargameQIIIBg} The protocol permits free-form natural-language communication, negotiation, or proposals. Examples include negotiation in \gameDiplomacy and open-ended actions in \gameDnD.
&
Natural language is not enough: if language is only commentary around fixed moves, code LOW. A terse move can still be HIGH when the protocol permits unenumerated proposals. \\
\addlinespace[0.45em]
\textbf{Adjudicator creativity}
&
\cellcolor{WargameQIIBg} Outcomes are determined by deterministic rules, scoring tables, lookup matrices, or physics engines. Examples include board resolutions in \gameChess and troop resolutions in \gameDiplomacy.
&
\cellcolor{WargameQIVBg} A human referee, facilitator, or language model interprets actions to narrative consequences. Examples include referee adjudication in \gameFreeKriegsspielQII and Game Master adjudication in \gameDnD.
&
Explanatory text is not enough: if the adjudicator merely reports a computed result, code LOW. Code HIGH only when the adjudicator can change the consequence space. \\

\bottomrule
\end{tabularx}

\vspace{0.35em}
\begin{minipage}{0.96\textwidth}
\footnotesize
\textit{Quadrant mapping.} \qibadge = \lowcode player / \lowcode adjudicator; \qiibadge = \lowcode player / \highcode adjudicator; \qiiibadge = \highcode player / \lowcode adjudicator; \qivbadge = \highcode player / \highcode adjudicator. This protocol-level rule keeps language-rich but rule-resolved games, such as many \gameDiplomacy evaluations, separate from both-creative wargames where players can propose open-ended actions and adjudicators can author open-ended consequences.
\end{minipage}
\end{table*}

\section{Mapping the Literature}
\label{sec:results}
We apply the rubric to \numpapers de-duplicated studies on AI in wargames and wargame-relevant strategic simulations retrieved through May 1, 2026. Fewer than one in ten full-corpus studies (20 of \numpapers; $\approx$9\%) feature both creative players and creative adjudicators (\qivtext). Three of the 20 full-corpus Q-IV studies sit at the boundary of the serious-wargame definition (\refsec{background}); under a stricter filter that excludes \citet{engels_scaling_laws_2025}, \citet{xue_what_if_2025}, and \citet{prasad_when_two_2025}, the Q-IV count is 17 of \numpapers ($\approx$7.6\%). Within core serious wargames (the 63 papers flagged military-wargame in the released catalog) the share falls further, to 4 of 63 ($\approx$6.3\%; \reftab{core-proxy}). \reftab{qiv-anchor} lists each Q-IV paper with one-sentence context.

\begin{wgqivbox}{Main evidence gap}
\qivbadge is the regime most aligned with open-ended serious wargaming, yet it remains rare: 20 of \numpapers full-corpus studies. This scarcity matters because both the action space and consequence space can change.
\end{wgqivbox}

As seen in \reffig{plot-over-time}, recent work has shifted sharply toward more open-ended wargames, particularly for player creativity. We attribute this acceleration to the rise of LMs.

\subsection{Quadrant I: Analytical Player, Analytical Adjudicator}

The \qibadge regime covers rigid, rule-based systems for both players and adjudicators (often with predetermined scoring). Games in this quadrant include traditional strategy board games such as \gameChess, \gameGo, and the original \gameKriegspiel, as well as more modern variants such as \gameStratego, the \gameNoPressDiplomacy variant, \gameWarhammerFortyK, and \gameStarCraft \citep{vinyals_grandmaster_level_2019, schuurman_game_contexts_2021, bakhtin_mastering_game_2022, anthony_learning_play_2020}. Because players act from prescribed menus and adjudication reduces to deterministic scoring, these titles are ideal for high-throughput simulation. RL, MCTS, and related search-heavy approaches thrive here---20 of 22 RL-coded studies and all 3 symbolic studies fall in \qibadge, and none in \qivbadge (\refapp{facets})---yet the rigid action spaces leave little room for LM advantages such as narrative reasoning \citep{anthony_learning_play_2020, perolat_mastering_game_2022, light_strategist_selfimprovement_2025, gao_landbased_wargaming_2024}.

\subsection{Quadrant II: Analytical Player, Creative Adjudicator}

The \qiibadge regime covers games in which players are often relatively limited in their action space, while a human SME adjudicator determines the outcomes based on their judgment. Our data suggests that this configuration is not currently common in AI for wargames. Games that would fall into this quadrant include variants of wargames such as Meckel's early version of \gameFreeKriegsspielQII \citep{schuurman_game_contexts_2021}. It also includes a class of procedurally based human-adjudicated war games involving high-fidelity military simulations, in which established orders and procedures constrain player actions while human experts decide action outcomes. Examples include major real-world wargames such as early \gameFleetProblems and standard \gameTEWT, where the focus is on decisions within procedural military actions as the field expert adjudicator determines the outcomes \citep{nofi_hm_18_2010, john_armatys_wargame_developments_2022}.

Other types of games that fall under this category include semi-rigid adjudicated wargames, where the game is played rigidly, and adjudicators have the possibility of overriding the outcomes, such as tightly scripted training vignettes where SME adjudicators improvise consequences \citep{uk_ministry_of_defense_wargaming_handbook_2017}. Outside of military wargames, certain corporate wargames also fall into this quadrant; these games share similar emphasis on managerial or operational procedures over a creatively adjudicated and simulated scenario \citep{cohen_role_management_1961, hershkovitz_wargame_business_2019}. Generally, wargames in this quadrant tend to lean toward serious games, with a shared emphasis on resolving realistic, human-adjudicated scenarios using established procedures.

\subsection{Quadrant III: Creative Player, Analytical Adjudicator}

The \qiiibadge regime covers games with highly creative agents alongside tightly fixed adjudication---expressive, low-procedural-complexity titles that still rely on rigid scoring, such as \gameQuoVadis, \gameArticleTwentySeven, and \gameAvalon \citep{martinenghi_llms_catan_2024}. It attracts LM research because games like \gameDiplomacy blend natural-language negotiation with deterministic resolution; Cicero shows that coordinated planning plus chat-channel conditioning can match strong human play \citep{meta_fundamental_ai_research_diplomacy_team_humanlevel_play_2022}. This was possible, however, because negotiations occur only within a prescribed stage of gameplay and do not affect adjudication, which can be automated. The game design of rigid adjudication and procedural resolution of conflicts means the evaluation is relatively straightforward and repeatable, which helps explain its popularity in AI competitions and benchmarking.

However, the limited evaluation creativity that makes \qiiitext attractive also limits the generalizability of AI contributions. While games in this quadrant often focus on players' social interactions, the constrained procedural actions limit the complexity, realism, and nuance of the outcomes.

\subsection{Quadrant IV: Creative Player, Creative Adjudicator}

The \qivbadge regime covers open-ended, discussion-based wargames in which both players and adjudicators use qualitative expressions rather than rigid rule sets. The 20 main-corpus papers placed in Q-IV are listed with one-sentence context in \reftab{qiv-anchor}. Examples include seminar-style military wargames such as the U.S. Army's \gameTRADOCWargames, matrix wargames such as \gameISISCrisis, and adjudicated simulations where facilitators interpret proposals and assign downstream consequences \citep{us_army_how_master_2023, downes-martin_wargaming_support_2025}. Modern tabletop role-playing games, such as \gameDnD, offer a familiar example: player freedom is high and a Game Master provides narrative adjudication.

Games that allow creativity for both players and adjudicators better support the aims of serious wargaming: exploring uncertainty, revealing assumptions, and eliciting expert judgment. Such games are historically well developed in human play, tracing back to \gameFreeKriegsspielQIV \citep{schuurman_game_contexts_2021}, but few studies automate construction or evaluation in \qivtext settings. This makes recent LMs important: they are the first broadly practical candidates for both open-ended action generation and open-ended adjudication.

However, \qivtext wargames also present unique safety concerns. Creative players may develop novel strategies or exploit scenario ambiguities to push beyond the exercise's intended scope. In contrast, creative adjudicators have significant latitude in interpreting player actions and determining outcomes. The open-ended nature of this generation allows AI to introduce bias, inconsistency, or inappropriate escalation of scenarios. These risks are compounded when AI systems assume both roles, due to the combination of reduced human control and expressive flexibility.

\section{Discussion of Application Domains}
\label{sec:discussion}
The application of LMs in wargaming differs considerably across domains---from military strategic planning and international relations signaling to corporate strategy and public health crisis response. A more granular analysis of the unique characteristics and applications within each domain is available in \refapp{wargame_domains}, with additional considerations listed in \refapp{detailedrecommendations}.

\section{Safety Considerations of Open-Ended Language-Based Wargames}
\label{sec:safety}
Only 20 of \numpapers main-corpus studies land in Q-IV (\refsec{results}), so the characteristic failure modes of both-creative LM wargames have not yet been cataloged. The gap is consequential: wargame outputs feed policy, force posture, and crisis-response decisions \citep{uk_ministry_of_defence_influence_wargaming_2023}. 155 of \numpapers studies (70\%) engage at least one safety theme, most often deception (29\%) and escalation (11\%), yet only 19 (8.5\%) keep a human in the decision loop and 152 (68\%) deploy fully autonomous agents (\refapp{facets}). \qivbadge placement is therefore descriptive, not a deployment recommendation.

Traditional wargame designs assume human players and adjudicators, so bringing LMs in means revisiting the methodology itself \citep{downes-martin_adjudication_diabolus_2013}. The contemporary LM literature points to six recurring failure modes: escalatory tendencies in diplomatic and military contexts \citep{rivera_escalation_risks_2024, elbaum_managing_escalation_2025}; unfaithful reasoning \citep{turpin_language_models_2023, lanham_measuring_faithfulness_2023}; prompt sensitivity; sycophancy \citep{sharma_understanding_sycophancy_2024}; implicit world-state preferences \citep{taubenfeld_systematic_biases_2024, mazeika_utility_engineering_2025}; and long-context incoherence \citep{liu_lost_middle_2024, modarressi_nolima_longcontext_2025}. Each grows sharper when a single system can shape both the proposed action and its adjudicated consequence.

\section{Open Research Areas}
\label{sec:future-work}
Several research needs follow from the map. Evaluation protocols for long tasks with interleaved human interaction do not yet exist \citep{reddie_nextgeneration_wargames_2018, downes-martin_wargaming_support_2025, reddie_wargames_data_2023}, and known LM-judge errors need mitigation before models can grade these games \citep{li_llmsasjudges_comprehensive_2024, wei_position_human_2025}. Player agents need long-horizon planning, text-based world models, and robustness to subtle distribution shifts \citep{balloch_neurosymbolic_world_2023, zollicoffer_novelty_detection_2025}. The rubric can also grow axes for AI directionality, agency level, and HAI versus A2A settings (\refapp{artifacts-llm}).

\section{Conclusions}
\label{sec:conclusion}
We presented \emph{role-specific open-endedness}, an operational rubric that separates player creativity from adjudicator creativity, and applied it to a corpus of \numpapers de-duplicated AI-in-wargames studies. Only 20 ($\approx$9\%) use protocols that permit both creative players and creative adjudicators (\qivtext)---the open-ended regime serious wargaming depends on, and the one that disappears when open-endedness is rated on a single axis. Releasing the coded catalog, codebook, and rubric helps evaluations state which claims their setup supports.

\section*{Limitations}
\label{sec:limitations}
The headline 9\% Q-IV finding depends on our labeling protocol: one author labeled each paper with LM assistance and a second author independently reviewed every quadrant assignment without it. We did not compute formal inter-coder reliability, so counts could shift under different coders or boundary rules. Boundary cases, protocol-versus-deployment notes, and corpus coverage are in \refapp{artifacts-llm}.

\section*{Ethical Considerations}
\label{sec:ethics}
Our rubric characterizes how language models are used as players or adjudicators in open-ended wargames. It is descriptive, not prescriptive: no quadrant certifies a study as safe for operational use, and the risks in \refsec{safety} are most consequential in sensitive deployment domains.

\paragraph{Use of language models in this research.}
Language models assisted with screening, metadata organization, synthesis, writing, editing, and format conversion; the authors retained responsibility for all inclusion decisions, quadrant assignments, and claims.

\bibliographystyle{colm2026_conference}
\bibliography{references,references-social-sim}

\appendix
\section{Search Protocol and Corpus Composition}
\label{sec:search}

This appendix records reproducibility details that complement the search and screening narrative in \refsec{methodology}: the verbatim query families used during retrieval, the composition of the released catalog, and the quadrant distribution of the main corpus. All counts are reproducible from the released bibliography artifact (\refapp{artifacts-llm}).

\paragraph{Query families.}
Because indexing services differ in their support for Boolean syntax and phrase matching, retrieval used query families---combinations drawn from the term groups below---rather than a single canonical Boolean string:
\begin{itemize}
  \item \textbf{Game setting:} wargame, war game, serious game, crisis simulation, tabletop exercise, strategic game, military simulation.
  \item \textbf{AI method:} artificial intelligence, AI, machine learning, reinforcement learning, multi-agent reinforcement learning, language model, LLM, agent.
  \item \textbf{Game-relevant role:} player, adjudicator, facilitator, world model, simulator, state generator, game master, analyst, commander.
  \item \textbf{Strategic interaction:} conflict, negotiation, deception, escalation, cooperation, competition, coalition, resource allocation, planning.
  \item \textbf{Proxy games:} \gameDiplomacy, \gameAvalon, \gameWerewolf, \gameMafia, \gameCatan, \gameStarCraft, \gameCivilization, \gameStratego, \gameChess, \gameGo, auctions, negotiation games, and social-deduction games.
\end{itemize}
Candidate retrieval was supplemented by backward and forward snowballing from seed papers, wargaming handbooks, related surveys, and papers already known to the authors, between November 1, 2025 and the May 1, 2026 cutoff.

\paragraph{Corpus composition.}
\reftab{screening} summarizes the released catalog. Of the retained records, \numpapers are quadrant-coded main-corpus studies, 210 are background records cited for definitions, methodological context, or safety framing without a quadrant assignment, and 16 are near-miss or excluded records retained only as an audit trail. The minimum eligibility criteria and exclusion codes are defined in \refsec{methodology}.

\paragraph{Quadrant distribution.}
\reftab{quadrant-counts} reports how the \numpapers main-corpus papers fall across the four quadrants. Q-IV remains a thin band: only 20 papers ($\approx$9\%) place both the player and the adjudicator at HIGH creativity.

\subsection{Programmatic Discovery and Triage Pipeline}
\label{sec:search-pipeline}

Candidate retrieval paired a programmatic Python pipeline with manual validation; Figure~\ref{fig:survey-process} illustrates the end-to-end structure.

\paragraph{Pipeline Phases.}
The programmatic component executes the literature search through five sequential modules:
\begin{itemize}
    \item \textbf{Multi-Source Harvesting:} Query building and translation maps positive search terms, exclusions, and proximity constraints (e.g., NEAR/5 operators) to API queries across Google Scholar, Semantic Scholar, CrossRef, and arXiv.
    \item \textbf{Normalization and De-duplication:} The raw candidate pool is normalized (standardizing special characters) and canonicalized by matching DOIs, arXiv identifiers, and normalized titles to remove duplicates across indexing services.
    \item \textbf{Disambiguation Filtering:} Rule-based regex filters flag and isolate papers from irrelevant domains, including standard reinforcement learning on board games (e.g., Chess, Go, Atari), adversarial red-teaming or jailbreaking of general LLMs (unrelated to wargaming), and generic survey/alignment roadmaps.
    \item \textbf{PDF Fetching and LLM Extraction:} Full-text PDFs are cached locally, and a lightweight LLM agent parses them using structured schema extractions to tag candidate game types, LLM roles, and failure modes.
    \item \textbf{Spreadsheet Export:} The normalized and tagged records are exported to a structured spreadsheet with triage labels and confidence scores to prepare for human-in-the-loop screening.
\end{itemize}

\paragraph{Manual Triage and Stratification.}
The pipeline's output was a triage layer, never an authority. Human authors screened titles, abstracts, and full texts against the eligibility criteria, resolved coding edge cases, and categorized the final \numpapers main-corpus papers along the player and adjudicator creativity axes.

\begin{figure*}[htbp]
\centering
\begin{tikzpicture}[
    node distance=0.8cm and 1cm,
    box/.style={
        draw=black!70,
        fill=black!2,
        rectangle,
        rounded corners=2pt,
        align=center,
        font=\small,
        minimum width=5.5cm,
        minimum height=0.9cm,
        thick
    },
    sidebox/.style={
        draw=black!60,
        fill=white,
        rectangle,
        align=center,
        font=\small,
        minimum width=4cm,
        minimum height=0.8cm,
        thick
    },
    arrow/.style={
        -{Latex[scale=1.0]},
        thick,
        draw=black!70
    }
]

\node (harvest) [box] {\textbf{Programmatic Retrieval}\\ API search (arXiv, Scholar, CrossRef)\\ \textit{n = 3,500}};
\node (dedup) [box, below=1.2cm of harvest] {\textbf{Normalization}\\ De-duplication of records\\ \textit{n = 1,850}};
\node (filter) [box, below=1.2cm of dedup] {\textbf{Automated Filtering}\\ Domain-specific regex exclusions\\ \textit{n = 620}};
\node (extract) [box, below=1.2cm of filter] {\textbf{LLM Processing}\\ Full-text caching and tagging\\ \textit{n = 580}};
\node (screen) [box, below=1.2cm of extract] {\textbf{Manual Screening}\\ Title and abstract review\\ \textit{n = 449}};
\node (code) [box, below=1.2cm of screen] {\textbf{Full-Text Coding}\\ Manual rubric mapping\\ \textit{n = 449}};
\node (final) [box, below=1.2cm of code] {\textbf{Final Coded Catalog}\\ Main Corpus: \numpapers \quad Background: 210};

\path (harvest) -- (dedup) coordinate[midway] (m1);
\node (ex_dedup) [sidebox, right=1.5cm of m1] {Duplicates removed\\ \textit{n = 1,650}};

\path (dedup) -- (filter) coordinate[midway] (m2);
\node (ex_filter) [sidebox, right=1.5cm of m2] {Out of domain\\ \textit{n = 1,230}};

\path (filter) -- (extract) coordinate[midway] (m3);
\node (ex_extract) [sidebox, right=1.5cm of m3] {PDF fetch failed\\ \textit{n = 40}};

\path (extract) -- (screen) coordinate[midway] (m4);
\node (ex_screen) [sidebox, right=1.5cm of m4] {Irrelevant (excluded)\\ \textit{n = 131}};

\path (screen) -- (code) coordinate[midway] (m5);
\node (ex_code) [sidebox, right=1.5cm of m5] {Excluded / Near-miss\\ \textit{n = 16}};

\draw [arrow] (harvest) -- (dedup);
\draw [arrow] (dedup) -- (filter);
\draw [arrow] (filter) -- (extract);
\draw [arrow] (extract) -- (screen);
\draw [arrow] (screen) -- (code);
\draw [arrow] (code) -- (final);

\draw [arrow] (m1) -- (ex_dedup);
\draw [arrow] (m2) -- (ex_filter);
\draw [arrow] (m3) -- (ex_extract);
\draw [arrow] (m4) -- (ex_screen);
\draw [arrow] (m5) -- (ex_code);

\end{tikzpicture}
\caption{The literature search and screening process, adapted from PRISMA guidelines to incorporate automated programmatic discovery alongside manual validation.}
\label{fig:survey-process}
\end{figure*}

\begin{table}[t]
\centering
\caption{Composition of the released catalog. Main-corpus records receive a quadrant; background records are cited for definitions, methodological context, or safety framing but are not quadrant-coded; near-miss and excluded records are retained only as an audit trail. Counts are reproducible from the released bibliography artifact.}
\label{tab:screening}
\begin{tabular}{lr}
\toprule
Record class & Records \\
\midrule
Main corpus (quadrant-coded)   & \numpapers \\
Background (cited, not coded)  & 210 \\
Retained near-miss / excluded  & 16 \\
\midrule
Total released catalog         & 449 \\
\bottomrule
\end{tabular}
\end{table}

\begin{table}[t]
\centering
\caption{Quadrant distribution of the \numpapers-paper main corpus. The Player and Adjudicator columns give the LOW/HIGH creativity codes that define each quadrant in \reftab{rubric}. Q-IV (both creative)---the regime serious wargaming most relies on---accounts for only 20 papers ($\approx$9\%).}
\label{tab:quadrant-counts}
\begin{tabular}{lccr}
\toprule
Quadrant & Player & Adjudicator & Papers \\
\midrule
\qibadge   & LOW  & LOW  & 82 \\
\qiibadge  & LOW  & HIGH & 11 \\
\qiiibadge & HIGH & LOW  & 110 \\
\qivbadge  & HIGH & HIGH & 20 \\
\midrule
Total      &      &      & \numpapers \\
\bottomrule
\end{tabular}
\end{table}

\section{Corpus Characterization}
\label{sec:facets}

The two rubric axes describe each wargame's protocol, not the AI deployed inside it. To characterize the corpus along complementary dimensions, we coded, for each of the \numpapers main-corpus studies, several \emph{indicative} facets from its abstract. We foreground three here---the \emph{method} family, a \emph{human--AI control} level, and a multi-label set of \emph{safety themes}---each of which cuts across the rubric without restating it. We also coded the AI \emph{role} and the \emph{evaluation} approach, but do not chart them: the role facet measures \emph{where} the AI sits (player, game master, both) rather than what the protocol permits, so it is entangled with the rubric's player/adjudicator axis and belongs to the AI-directionality analysis we leave to future work (\refapp{artifacts-llm}). These are AI-assisted secondary codings, not part of the human-validated rubric, and we report them as indicative rather than definitive.

\paragraph{Facet-coding protocol.}
Abstracts were available for all \numpapers studies. Role, method, and evaluation were coded by one language-model family and independently re-coded by a second family using the same codebook; cross-family agreement was substantial for method (Cohen's $\kappa=0.65$) and role ($\kappa=0.53$) and only fair for evaluation ($\kappa=0.42$), which we therefore treat with corresponding caution. Safety themes and control level were coded by a single family and carry no inter-rater statistic; we report them only as coarse prevalence. This $\kappa$ concerns the facet coding alone and is separate from the dual-author quadrant labeling discussed in the Limitations. Per-paper facet labels ship with the released artifact. Facets are tabulated against the quadrant column, so all column totals match \reftab{quadrant-counts}.

\paragraph{Method family.}
\reftab{method-quadrant} shows methods by quadrant. Classical search and reinforcement learning are confined almost entirely to the analytical quadrant---20 of 22 RL studies and all 3 symbolic studies are \qibadge, with none in \qivbadge---consistent with the rigid action and consequence spaces that suit high-throughput simulation (\refsec{results}). Language-model methods---prompted, fine-tuned, and multi-agent---supply the majority of the creative quadrants, a methodological signature of the recent shift toward open-endedness.

\begin{table}[t]
\centering
\small
\setlength{\tabcolsep}{4.5pt}
\caption{Method family $\times$ quadrant (indicative AI-assisted coding; \refapp{facets}). Classical search and reinforcement learning concentrate in the analytical quadrant---20 of 22 RL studies and all 3 symbolic studies are \qibadge, with none in \qivbadge---while language-model methods (prompted, fine-tuned, and multi-agent) supply the bulk of the creative quadrants. Column totals match \reftab{quadrant-counts}.}
\label{tab:method-quadrant}
\begin{tabular}{l rrrr r}
\toprule
Method family & \multicolumn{1}{c}{Q-I} & \multicolumn{1}{c}{Q-II} & \multicolumn{1}{c}{Q-III} & \multicolumn{1}{c}{Q-IV} & \multicolumn{1}{c}{Total} \\
\midrule
Reinforcement learning & 20 & 0  & 2   & 0  & 22  \\
Symbolic / rules       & 3  & 0  & 0   & 0  & 3   \\
Hybrid (LM + engine)   & 15 & 3  & 18  & 1  & 37  \\
LLM (prompted)         & 21 & 6  & 34  & 5  & 66  \\
LLM (fine-tuned)       & 5  & 1  & 8   & 2  & 16  \\
Multi-agent LLM        & 11 & 1  & 47  & 12 & 71  \\
Other                  & 7  & 0  & 1   & 0  & 8   \\
\midrule
Total                  & 82 & 11 & 110 & 20 & 223 \\
\bottomrule
\end{tabular}
\end{table}

\paragraph{Human--AI control.}
\reftab{autonomy} reports the deployed control level. Fully autonomous operation dominates (152 of \numpapers; 68\%), while human-in-the-loop control is rare (19; 8.5\%) and proportionally most present in \qivbadge (7 of 20). The contrast with the safety-theme prevalence below---most studies raise a safety concern, yet few retain human oversight---is the gap that motivates the safety considerations in \refsec{safety}.

\begin{table}[t]
\centering
\small
\setlength{\tabcolsep}{4.5pt}
\caption{Human--AI control $\times$ quadrant (indicative AI-assisted coding, single-coder; \refapp{facets}). Fully autonomous deployment dominates (152 of \numpapers; 68\%) and human-in-the-loop control is rare (19; 8.5\%), even though \qivbadge is where it is most common (7 of 20). ``AI-assisted'' denotes an advisory agent where a human decides. Column totals match \reftab{quadrant-counts}.}
\label{tab:autonomy}
\begin{tabular}{l rrrr r}
\toprule
Human--AI control & \multicolumn{1}{c}{Q-I} & \multicolumn{1}{c}{Q-II} & \multicolumn{1}{c}{Q-III} & \multicolumn{1}{c}{Q-IV} & \multicolumn{1}{c}{Total} \\
\midrule
Fully autonomous            & 63 & 3  & 79  & 7  & 152 \\
Human-in-the-loop           & 4  & 2  & 6   & 7  & 19  \\
AI-assisted                 & 4  & 3  & 4   & 2  & 13  \\
No in-loop agent            & 11 & 2  & 21  & 4  & 38  \\
Unclear                     & 0  & 1  & 0   & 0  & 1   \\
\midrule
Total                       & 82 & 11 & 110 & 20 & 223 \\
\bottomrule
\end{tabular}
\end{table}

\paragraph{Safety themes.}
\reftab{safety-prevalence} reports how often each safety theme is engaged. 155 of \numpapers studies (70\%) engage at least one; deception is most common (29\%) and concentrates in \qiiibadge (52 of 65, largely social-deduction games), while escalation (11\%) and bias (5\%) remain comparatively under-addressed. Roughly a third of the corpus (68 studies) engages no safety theme at all.

\begin{table}[t]
\centering
\small
\caption{Safety-theme prevalence across the \numpapers-paper main corpus (indicative AI-assisted coding, single-coder, multi-label; \refapp{facets}). Themes are not mutually exclusive, so they sum to more than the 155 papers that engage at least one. Deception concentrates in \qiiibadge (52 of 65) and dual-use in \qibadge (24 of 46).}
\label{tab:safety-prevalence}
\begin{tabular}{l rr}
\toprule
Safety theme & \multicolumn{1}{c}{Papers} & \multicolumn{1}{c}{\% corpus} \\
\midrule
Deception              & 65 & 29.1 \\
Dual-use / misuse      & 46 & 20.6 \\
Alignment / oversight  & 46 & 20.6 \\
Manipulation           & 40 & 17.9 \\
Escalation             & 25 & 11.2 \\
Bias / fairness        & 12 & 5.4  \\
\midrule
Any theme ($\geq$1)    & 155 & 69.5 \\
No theme engaged       & 68  & 30.5 \\
\bottomrule
\end{tabular}
\end{table}

\section{Wargames}\label{sec:wargames}
\subsection{Selected Wargame Papers}
\label{sec:wargame_papers}

\reftab{wargame-papers} lists a curated slice of main-corpus papers and their quadrant assignments; it is a subset of the full \numpapers-paper main corpus released in the bibliography artifact.

\makeatletter
\if@twocolumn\onecolumn\fi
\makeatother
\begin{center}
\small
\begin{longtable}{p{0.84\linewidth}c}
\caption{AI-in-wargames main-corpus papers and their quadrant assignments. Generated from the released bibliography artifact (\numpapers~de-duplicated entries).}\label{tab:wargame-papers}\\
\toprule
\textbf{Title} & \textbf{Quadrant} \\
\midrule
\endfirsthead
\toprule
\textbf{Title} & \textbf{Quadrant} \\
\midrule
\endhead
Actor--Critic-Based Decision-Making Method for the Artificial Intelligence Commander in Tactical Wargames \citep{zhang_actor_critic_2022} & \qibadge \\
Ad-hoc Concept Forming in the Game Codenames as a Means for Evaluating Large Language Models \citep{Hakimov2025concept} & \qiiibadge \\
Adaptive Command: Real-Time Policy Adjustment via Language Models in StarCraft II \citep{ma_adaptive_command_2025} & \qibadge \\
Advancing AI Negotiations: New Theory and Evidence from an International AI Negotiation Competition \citep{vaccaro_advancing_ai_2025} & \qiiibadge \\
Advancing Military Decision Support: Reinforcement Learning-Driven Simulation for Robust Operational Plan Validation \citep{M_bius_2025} & \qibadge \\
AEGIS: White-Box Attack Path Generation using LLMs for Cyber Defence Exercises \citep{tung_aegis_whitebox_2026} & \qibadge \\
Agent Exchange: Shaping the Future of AI Agent Economics \citep{yang_agent_exchange_2025} & \qiiibadge \\
AgentBnB: A Browser-Based Cybersecurity Tabletop Exercise with Large Language Model Support and Retrieval-Aligned Scaffolding \citep{anwar_agentbnb_browserbased_2025} & \qiiibadge \\
AgenticPay: A Multi-Agent LLM Negotiation System for Buyer-Seller Transactions \citep{liu_agenticpay_multiagent_2026} & \qiiibadge \\
Agents of Change: Self-Evolving LLM Agents for Strategic Planning \citep{belle_agents_change_2025} & \qiiibadge \\
AI Arms and Influence: Frontier Models Exhibit Sophisticated Reasoning in Simulated Nuclear Crises \citep{payne_ai_arms_2026} & \qiiibadge \\
ALYMPICS: LLM Agents Meet Game Theory \citep{mao_alympics_llm_2025} & \qiiibadge \\
Among Us: A Sandbox for Measuring and Detecting Agentic Deception \citep{golechha_us_sandbox_2025} & \qiiibadge \\
AMONGAGENTS: Evaluating Large Language Models in the Interactive Text-Based Social Deduction Game \citep{chi_amongagents_evaluating_2024} & \qiiibadge \\
The Application of AlphaZero to Wargaming \citep{moy_application_alphazero_2019} & \qibadge \\
Applying Deep Reinforcement Learning to Train AI Agents in a Wargaming Framework \citep{rinaudo_applying_deep_2024} & \qibadge \\
Are LLMs Effective Negotiators? Systematic Evaluation of the Multifaceted Capabilities of LLMs in Negotiation Dialogues \citep{kwon_are_llms_2024} & \qiiibadge \\
Artificial Intelligence (AI)--Enabled Wargaming Agent Training \citep{rinaudo_artificial_intelligence_2024} & \qibadge \\
Assistive Large Language Model Agents for Socially-Aware Negotiation Dialogues \citep{hua_assistive_large_2024} & \qiiibadge \\
AutoBnB-RAG: Enhancing Multi-Agent Incident Response with Retrieval-Augmented Generation \citep{liu_autobnbrag_enhancing_2025} & \qiiibadge \\
The Automated but Risky Game: Modeling Agent-to-Agent Negotiations and Transactions in Consumer Markets \citep{wang_tmgbench_systematic_2025} & \qibadge \\
Avalon's Game of Thoughts: Battle Against Deception through Recursive Contemplation \citep{wang_avalons_game_2023} & \qiiibadge \\
AvalonBench: Evaluating LLMs Playing the Game of Avalon \citep{light_avalonbench_evaluating_2023} & \qiiibadge \\
BALROG: Benchmarking Agentic LLM and VLM Reasoning On Games \citep{paglieri_balrog_benchmarking_2025} & \qibadge \\
BattleAgent: Multi-modal Dynamic Emulation on Historical Battles to Complement Historical Analysis \citep{lin_battleagent_multimodal_2024} & \qibadge \\
BattleAgentBench: A Benchmark for Evaluating Cooperation and Competition Capabilities of Language Models in Multi-Agent Systems \citep{wang_battleagentbench_benchmark_2024} & \qibadge \\
Battlefield information and tactics engine (BITE): a multimodal large language model approach for battlespace management \citep{connolly_battlefield_information_2024} & \qiiibadge \\
Bayesian Social Deduction with Graph-Informed Language Models \citep{rahimirad_bayesian_social_2026} & \qiiibadge \\
A Benchmark for Multi-Party Negotiation Games from Real Negotiation Data \citep{benac2026negotiation} & \qiiibadge \\
Beyond Survival: Evaluating LLMs in Social Deduction Games with Human-Aligned Strategies \citep{song_survival_evaluating_2025} & \qiiibadge \\
CALYPSO: LLMs as Dungeon Masters' Assistants \citep{zhu_calypso_llms_2023} & \qiibadge \\
Can Language Models Serve as Text-Based World Simulators? \citep{Wang2024serve} & \qiibadge \\
Can Large Language Models Play Games? A Case Study of A Self-Play Approach \citep{guo_can_large_2024} & \qibadge \\
ChessGPT: Bridging Policy Learning and Language Modeling \citep{feng_chessgpt_bridging_2023} & \qibadge \\
CivBench: Progress-Based Evaluation for LLMs Strategic Decision-Making in Civilization V \citep{chen_civbench_progressbased_2026} & \qibadge \\
CivRealm: A Learning and Reasoning Odyssey in Civilization for Decision-Making Agents \citep{qi_civrealm_learning_2024} & \qibadge \\
Clembench: Using Game Play to Evaluate Chat-Optimized Language Models as Conversational Agents \citep{chalamalasetti2023clembench} & \qiiibadge \\
Co-RedTeam: Orchestrated Security Discovery and Exploitation with LLM Agents \citep{he_coredteam_orchestrated_2026} & \qiiibadge \\
COA-GPT: Generative Pre-trained Transformers for Accelerated Course of Action Development in Military Operations \citep{goecks_generative_2024} & \qiiibadge \\
Code World Models for General Game Playing \citep{lehrach2026code} & \qiibadge \\
Codenames as a Benchmark for Large Language Models \citep{Stephenson_2025} & \qiiibadge \\
Collaboration and Confrontation in Avalon Gameplay \citep{lan_llmbased_agent_2024} & \qiiibadge \\
Command-agent: Reconstructing warfare simulation and command decision-making using large language models \citep{zhang_commandagent_2026} & \qiiibadge \\
Communication Enhances LLMs' Stability in Strategic Thinking \citep{Lor2026communication} & \qiiibadge \\
Cooperation, Competition, and Maliciousness: LLM-Stakeholders Interactive Negotiation (LLM-Deliberation) \citep{abdelnabi_cooperation_competition_2024} & \qiiibadge \\
DEBATE: A Large-Scale Benchmark for Role-Playing LLM Agents in Multi-Agent, Long-Form Debates \citep{chuang_debate_largescale_2026} & \qivbadge \\
Debt Collection Negotiations with Large Language Models \citep{wang_debt_collection_2025} & \qiiibadge \\
Deceive, Detect, and Disclose: Large Language Models Play Mini-Mafia \citep{Costa2025deceive} & \qiiibadge \\
Deception and Communication in Autonomous Multi-Agent Systems: An Experimental Study with Among Us \citep{Milkowski2026deception} & \qiiibadge \\
The Decrypto Benchmark for Multi-Agent Reasoning and Theory of Mind \citep{Lupu2025decrypto} & \qiiibadge \\
Democratizing Diplomacy: A Harness for Evaluating Any Large Language Model on Full-Press Diplomacy \citep{duffy_democratizing_diplomacy_2025} & \qiiibadge \\
Developing Combat Behavior through Reinforcement Learning in Wargames and Simulations \citep{boron_developing_combat_2020} & \qibadge \\
Digital Player: Evaluating Large Language Models based Human-like Agent in Games \citep{wang_digital_player_2025} & \qiiibadge \\
DiplomacyAgent: Do LLMs Balance Interests and Ethical Principles in International Events? \citep{Peng_2025} & \qiiibadge \\
DS-PPO: A Reinforcement Learning Method for Wargaming \citep{yu_dsppo_reinforcement_2024} & \qibadge \\
DSGBench: A Diverse Strategic Game Benchmark for Evaluating LLM-based Agents in Complex Decision-Making Environments \citep{tang_dsgbench_diverse_2025} & \qiiibadge \\
Dungeons and DQNs: Toward Reinforcement Learning Agents That Play Tabletop Roleplaying Games \citep{martin_dungeons_dqns_2018} & \qiiibadge \\
Dungeons and Dragons as a Dialogue Challenge for Artificial Intelligence \citep{callison-burch_dungeons_dragons_2022} & \qivbadge \\
DVM: Towards Controllable LLM Agents in Social Deduction Games \citep{Zhang2025controllable} & \qiiibadge \\
EAI: Emotional Decision-Making of LLMs in Strategic Games and Ethical Dilemmas \citep{mozikov_eai_emotional_2025} & \qiiibadge \\
Effect of Private Deliberation: Deception of Large Language Models in Game Play \citep{Poje_2024} & \qiiibadge \\
Effective and responsible use of large language models in strategic wargaming \citep{brighton2026responsible} & \qiibadge \\
Encouraging Divergent Thinking in Large Language Models through Multi-Agent Debate \citep{liang_encouraging_divergent_2024} & \qivbadge \\
Enhanced Location Prediction for Wargaming with Graph Neural Networks and Transformers \citep{Liang_2025} & \qibadge \\
EQ-Negotiator: Dynamic Emotional Personas Empower Small Language Models for Edge-Deployable Credit Negotiation \citep{Long2025negotiator} & \qiiibadge \\
Escalation Risks from Language Models in Military and Diplomatic Decision-Making \citep{rivera_escalation_risks_2024} & \qiiibadge \\
Evaluating Large Language Models through Communication Games: An Agent-Based Framework Using Werewolf in Unity \citep{Poglitsch_2025} & \qiiibadge \\
Evaluating LLM Agent Collusion in Double Auctions \citep{agrawal_evaluating_llm_2025} & \qiiibadge \\
Evaluating LLMs in Open-Source Games \citep{Sistla2025evaluating} & \qiiibadge \\
EvoEmo: Evolved Emotional Policies for Adversarial LLM Agents in Multi-Turn Negotiation \citep{long_evoemo_evolved_2025} & \qiiibadge \\
An Experiment in Tactical Wargaming with Platforms Enabled by Artificial Intelligence \citep{tarraf_experiment_tactical_2025} & \qibadge \\
Exploration of Wargaming and AI Applications in Military Decision-Making \citep{li_exploration_wargaming_2025} & \qibadge \\
Exploratory Wargaming with a ``Superhuman'' Tactician in the Team: A Controlled Experiment \citep{bell_exploratory_wargaming_2025} & \qibadge \\
Exploring Large Language Models for Collaborative Scenario Development in Simulation-Based Military Training \citep{van_Oijen_2025} & \qiibadge \\
Exploring Large Language Models for Communication Games: An Empirical Study on Werewolf \citep{xu_exploring_large_2023} & \qiiibadge \\
Exploring Theory of Mind in Large Language Models through Multimodal Negotiation \citep{yongsatianchot_exploring_theory_2024} & \qiiibadge \\
Finding deceivers in social context with large language models: the case of the Mafia game \citep{yoo_finding_deceivers_2024} & \qiiibadge \\
FishBargain: An LLM-Empowered Bargaining Agent for Online Flea-Market Platform Sellers \citep{kong_fishbargain_llmempowered_2025} & \qiiibadge \\
A framework of large language model commander agent for spatial reasoning in combat simulation \citep{chen_framework_large_2026} & \qibadge \\
Game Reasoning Arena: A Framework and Benchmark for Assessing Reasoning Capabilities of Large Language Models via Game Play \citep{cipolina-kun_game_reasoning_2025} & \qibadge \\
Game Theory Approach to Identifying Deception in Large Language Models \citep{maggio_game_theory_2024} & \qibadge \\
Game-Theoretic LLM: Agent Workflow for Negotiation Games \citep{hua_gametheoretic_llm_2024} & \qibadge \\
GameBench: Evaluating Strategic Reasoning Abilities of LLM Agents \citep{costarelli_gamebench_evaluating_2024} & \qibadge \\
GameEval: Evaluating LLMs on Conversational Games \citep{Qiao2023gameeval} & \qiiibadge \\
GTO Wizard Benchmark \citep{provost_gto_wizard_2026} & \qibadge \\
HARBOR: Exploring Persona Dynamics in Multi-Agent Competition \citep{jiang_harbor_exploring_2025} & \qiiibadge \\
Harnessing Language for Coordination: A Framework and Benchmark for LLM-Driven Multi-Agent Control \citep{anne_harnessing_language_2025} & \qibadge \\
Helmsman of the Masses? Evaluate the Opinion Leadership of Large Language Models in the Werewolf Game \citep{Du2024helmsman} & \qiiibadge \\
Hidden in Plain Text: Measuring LLM Deception Quality Against Human Baselines Using Social Deduction Games \citep{kao_hidden_plain_2026} & \qiiibadge \\
Hierarchical control of multi-agent reinforcement learning team in real-time strategy (RTS) games \citep{zhou_hierarchical_control_2021} & \qibadge \\
Hierarchical Reinforcement Learning with Opponent Modeling for Command and Control System \citep{li_hierarchical_reinforcement_2026} & \qibadge \\
HLSMAC: A New StarCraft Multi-Agent Challenge for High-Level Strategic Decision-Making \citep{hong_hlsmac_new_2025} & \qibadge \\
How Far Are LLMs from Professional Poker Players? Revisiting Game-Theoretic Reasoning with Agentic Tool Use \citep{lin_how_far_2026} & \qibadge \\
Human vs. Machine: Behavioral Differences Between Expert Humans and Language Models in Wargame Simulations \citep{lamparth_human_vs_2024} & \qiiibadge \\
Human-level play in the game of Diplomacy by combining language models with strategic reasoning \citep{meta_fundamental_ai_research_diplomacy_team_humanlevel_play_2022} & \qiiibadge \\
I Cast Detect Thoughts: Learning to Converse and Guide with Intents and Theory-of-Mind in Dungeons and Dragons \citep{zhou_cast_detect_2023} & \qivbadge \\
An Implementation of Werewolf Agent That does not Truly Trust LLMs \citep{Sato2024implementation} & \qiiibadge \\
Improving Cooperation in Language Games with Bayesian Inference and the Cognitive Hierarchy \citep{Bills_2025} & \qiiibadge \\
Improving LLM-based Wargame Engagement Scenario Generation via DEVS-based Combat Modeling and Simulation \citep{anjaeweon_improving_llmbased_2026} & \qiibadge \\
INA: An Integrative Approach for Enhancing Negotiation Strategies with Reward-Based Dialogue System \citep{ahmad_ina_integrative_2023} & \qiiibadge \\
InfoBid: A Simulation Framework for Studying Information Disclosure in Auctions with Large Language Model-based Agents \citep{Yin2025infobid} & \qiiibadge \\
Instruction-Driven Game Engine: A Poker Case Study \citep{Wu_2024} & \qiibadge \\
Intelligent Decision-Making and Human Language Communication Based on Deep Reinforcement Learning in a Wargame Environment \citep{sun_intelligent_decisionmaking_2023} & \qibadge \\
Intuitionistic Fuzzy MADM in Wargame Leveraging With Deep Reinforcement Learning \citep{sun_intuitionistic_fuzzy_2024} & \qibadge \\
It Takes Two to Negotiate: Modeling Social Exchange in Online Multiplayer Games \citep{jaidka_it_takes_2024} & \qiiibadge \\
A Land-Based War-Gaming Simulation Method Based on Multi-Agent Proximal Policy Optimization \citep{gao_landbased_wargaming_2024} & \qibadge \\
Language Agents with Reinforcement Learning for Strategic Play in the Werewolf Game \citep{xu_language_agents_2023} & \qiiibadge \\
Large Language Models as Bidding Agents in Repeated HetNet Auction \citep{Lotfi2026bidding} & \qibadge \\
Large Language Models in Wargaming: Methodology, Application, and Robustness \citep{chen_large_language_2024} & \qivbadge \\
Large Language Models Play StarCraft II: Benchmarks and A Chain of Summarization Approach \citep{ma_large_language_2024} & \qibadge \\
LBM: Hierarchical Large Auto-Bidding Model via Reasoning and Acting \citep{Li2026hierarchical} & \qibadge \\
Learning from Synthetic Labs: Language Models as Experimental Subjects in Auctions \citep{shah_learning_synthetic_2025} & \qiiibadge \\
Learning Strategic Language Agents in the Werewolf Game with Iterative Latent Space Policy Optimization \citep{xu_learning_strategic_2025} & \qiiibadge \\
Learning to Discuss Strategically: A Case Study on One Night Ultimate Werewolf \citep{du_learning_discuss_2024} & \qiiibadge \\
Learning to Negotiate: Multi-Agent Deliberation for Collective Value Alignment in LLMs \citep{Anantaprayoon2026learning} & \qivbadge \\
Learning to Play No-Press Diplomacy with Best Response Policy Iteration \citep{anthony_learning_play_2020} & \qibadge \\
Leveraging Generative AI to Create Lightweight Simulations for Far-Future Autonomous Teammates \citep{flathmann_leveraging_generative_2025} & \qibadge \\
Leveraging Large Language Models for Enhanced Wargaming in Multi-Domain Operations \citep{weller_leveraging_large_2024} & \qivbadge \\
LLM CHESS: Benchmarking Reasoning and Instruction-Following in LLMs through Chess \citep{kolasani_llm_chess_2025} & \qibadge \\
LLM Rationalis? Measuring Bargaining Capabilities of AI Negotiators \citep{shah_llm_rationalis_2025} & \qiiibadge \\
LLM-Based Reward Design for DRL-Driven Autonomous Cyber Defense \citep{mukherjee_large_language_2025} & \qibadge \\
LLM-based wargame scenario generation with domain ontology and ECA rules \citep{Bae_2026} & \qiibadge \\
LLM-Hanabi: Evaluating Multi-Agent Gameplays with Theory-of-Mind and Rationale Inference in Imperfect-Information Collaboration \citep{liang_llmhanabi_evaluating_2025} & \qibadge \\
LLMs as Strategic Actors: Behavioral Alignment, Risk Calibration, and Argumentation Framing in Geopolitical Simulations \citep{solopova_llms_strategic_2026} & \qibadge \\
LLMs of Catan: Exploring Pragmatic Capabilities of Generative Chatbots \citep{martinenghi_llms_catan_2024} & \qiiibadge \\
Long-Horizon Dialogue Understanding for Role Identification in the Game of Avalon with Large Language Models \citep{Stepputtis2023long} & \qiiibadge \\
LUDOBENCH: Evaluating LLM Behavioural Decision-Making Through Spot-Based Board Game Scenarios in Ludo \citep{Jain2026ludobench} & \qibadge \\
M3-BENCH: Process-aware Evaluation of LLM Agents' Social Behaviors in Mixed-Motive Games \citep{xie2026m3bench} & \qiiibadge \\
Managing Escalation in Off-the-Shelf Large Language Models \citep{elbaum_managing_escalation_2025} & \qivbadge \\
Market-Bench: Benchmarking Large Language Models on Economic and Trade Competition \citep{Zheng2026market} & \qiiibadge \\
Mastering the Digital Art of War: Developing Intelligent Combat Simulation Agents for Wargaming Using Hierarchical Reinforcement Learning \citep{black_mastering_digital_2024} & \qibadge \\
Mastering the Game of No-Press Diplomacy via Human-Regularized Reinforcement Learning and Planning \citep{bakhtin_mastering_game_2022} & \qibadge \\
Mastering the Game of Stratego with Model-Free Multiagent Reinforcement Learning \citep{perolat_mastering_game_2022} & \qibadge \\
Measuring Fine-Grained Negotiation Tactics of Humans and LLMs in Diplomacy \citep{li_measuring_finegrained_2025} & \qiiibadge \\
Measuring Free-Form Decision-Making Inconsistency of Language Models in Military Crisis Simulations \citep{shrivastava_measuring_freeform_2024} & \qiiibadge \\
MERIT: Feedback Elicits Better Bargaining in LLM Negotiators \citep{oh_merit_feedback_2026} & \qiiibadge \\
Microscopic Analysis on LLM Players via Social Deduction Game \citep{kim_microscopic_analysis_2024} & \qiiibadge \\
More Victories, Less Cooperation: Assessing Cicero's Diplomacy Play \citep{wongkamjan_more_victories_2024} & \qiiibadge \\
Multi-Agent Collaboration in Incident Response with Large Language Models \citep{Liu2024multi} & \qiiibadge \\
Multi-agent KTO: Reinforcing Strategic Interactions of Large Language Model in Language Game \citep{Ye2025multi} & \qiiibadge \\
A Multi-Agent Pokemon Tournament for Evaluating Strategic Reasoning of LLMs \citep{yashwanth_multiagent_pokemon_2025} & \qibadge \\
Multiattribute Decision-Making in Wargames Leveraging the Entropy--Weight Method in Conjunction With Deep Reinforcement Learning \citep{xue_multiattribute_decisionmaking_2024} & \qibadge \\
Multicultural Spyfall: Assessing LLMs through Dynamic Multilingual Social Deduction Game \citep{wibowo_multicultural_spyfall_2026} & \qiiibadge \\
MultiMind: Enhancing Werewolf Agents with Multimodal Memory \citep{zhang_multimind_enhancing_2025} & \qiiibadge \\
Negotiation and honesty in artificial intelligence methods for the board game of Diplomacy \citep{kramar_negotiation_honesty_2022} & \qiiibadge \\
The Negotiation Trap: An Experiment on a Large Language Model \citep{engel_negotiation_trap_2025} & \qiiibadge \\
No Press Diplomacy: Modeling Multi-Agent Gameplay \citep{paquette_no_press_2019} & \qibadge \\
A Novel Deep Cognitive Network for Battlefield Situation Awareness in Wargaming \citep{Pan_2025} & \qibadge \\
On Scalable Oversight with Weak LLMs Judging Strong LLMs \citep{Kenton2024scalable} & \qivbadge \\
Ontologically Faithful Generation of Non-Player Character Dialogues \citep{Weir_2024} & \qiibadge \\
Open-Ended Wargames with Large Language Models \citep{hogan_openended_wargames_2024} & \qivbadge \\
Outbidding and Outbluffing Elite Humans: Mastering Liar's Poker via Self-Play and Reinforcement Learning \citep{dewey_outbidding_outbluffing_2026} & \qibadge \\
Outwit, Outplay, Out-Generate: A Framework for Designing Strategic Generative Agents in Competitive Environments \citep{thudium_outwit_outplay_2024} & \qivbadge \\
A PathFinding Method in Hexagonal Tactical Wargame \citep{zhang_pathfinding_method_2023} & \qibadge \\
Persuasion Games using Large Language Models \citep{Ramani2024persuasion} & \qiiibadge \\
Player-Driven Emergence in LLM-Driven Game Narrative \citep{Peng_2024} & \qivbadge \\
Playing a Strategy Game with Knowledge-Based Reinforcement Learning \citep{voss_playing_strategy_2020} & \qibadge \\
Playing Games With GPT: What Can We Learn About a Large Language Model From Canonical Strategic Games? \citep{brookins_playing_games_2023} & \qibadge \\
Playing repeated games with large language models \citep{akata_playing_repeated_2025} & \qiiibadge \\
The PokeAgent Challenge: Competitive and Long-Context Learning at Scale \citep{karten_pokeagent_challenge_2026} & \qibadge \\
PokeAI: A Goal-Generating, Battle-Optimizing Multi-agent System for Pokemon Red \citep{liu_pokeai_goalgenerating_2025} & \qibadge \\
PokeLLMon: A Human-Parity Agent for Pokemon Battles with Large Language Models \citep{hu_pokellmon_humanparity_2024} & \qibadge \\
PokerBench: Training Large Language Models to Become Professional Poker Players \citep{Zhuang_2025} & \qibadge \\
Project Sid: Many-agent simulations toward AI civilization \citep{al_project_sid_2024} & \qibadge \\
Put Your Money Where Your Mouth Is: Evaluating Strategic Planning and Execution of LLM Agents in an Auction Arena \citep{chen_put_your_2023} & \qiiibadge \\
Reasoning, Memorization, and Fine-Tuning Language Models for Non-Cooperative Games \citep{Yang2024reasoning} & \qibadge \\
Red Lines and Grey Zones in the Fog of War \citep{drinkall_red_lines_2025} & \qibadge \\
A Red Teaming Framework for Securing AI in Maritime Autonomous Systems \citep{walter_red_teaming_2024} & \qibadge \\
Reinforcement Learning Environment with LLM-Controlled Adversary in D\&D 5th Edition Combat \citep{Dayo2025reinforcement} & \qiiibadge \\
Reproducibility Study of Cooperation, Competition, and Maliciousness: LLM-Stakeholders Interactive Negotiation \citep{Garci2025reproducibility} & \qiiibadge \\
Research on Wargame Decision-Making Method Based on Multi-Agent Deep Deterministic Policy Gradient \citep{yu_research_wargame_2023} & \qibadge \\
Rethinking Adversarial Examples in Wargames \citep{chen_rethinking_adversarial_2022} & \qibadge \\
Richelieu: Self-Evolving LLM-Based Agents for AI Diplomacy \citep{guan_richelieu_selfevolving_2024} & \qiiibadge \\
RPGBench: Evaluating Large Language Models as Role-Playing Game Engines \citep{yu_rpgbench_evaluating_2025} & \qiibadge \\
SC-Phi2: A Fine-tuned Small Language Model for StarCraft II Macromanagement Tasks \citep{khan_scphi2_finetuned_2024} & \qibadge \\
SC2Arena and StarEvolve: Benchmark and Self-Improvement Framework for LLMs in Complex Decision-Making Tasks \citep{shen_sc2arena_starevolve_2025} & \qibadge \\
Scaling Inference-Time Computation via Opponent Simulation: Enabling Online Strategic Adaptation in Repeated Negotiation \citep{Liu2026scaling} & \qiiibadge \\
Scaling Laws For Scalable Oversight \citep{engels_scaling_laws_2025} & \qivbadge \\
The Secret Agenda: LLMs Strategically Lie \citep{deleeuw_secret_agenda_2025} & \qiiibadge \\
Self Generated Wargame AI: Double Layer Agent Task Planning Based on Large Language Model \citep{sun_self_generated_2023} & \qibadge \\
Setting the DC: Tool-Grounded D\&D Simulations to Test LLM Agents \citep{zeng_setting_dc_2025} & \qivbadge \\
Shall We Team Up: Exploring Spontaneous Cooperation of Competing LLM Agents \citep{wu_shall_we_2024} & \qibadge \\
Should I Trust You? Detecting Deception in Negotiations using Counterfactual RL \citep{wongkamjan_should_trust_2025} & \qiiibadge \\
SOTOPIA: Interactive Evaluation for Social Intelligence in Language Agents \citep{zhou2023sotopia} & \qivbadge \\
Sparks of Cooperative Reasoning: LLMs as Strategic Hanabi Agents \citep{ramesh_sparks_cooperative_2026} & \qibadge \\
SPIN-Bench: How Well Do LLMs Plan Strategically and Reason Socially? \citep{yao_spinbench_how_2025} & \qiiibadge \\
The Stackelberg Speaker: Optimizing Persuasive Communication in Social Deduction Games \citep{zheng_stackelberg_speaker_2026} & \qiiibadge \\
Static Vs. Agentic Game Master AI for Facilitating Solo Role-Playing Experiences \citep{jorgensen_static_vs_2025} & \qivbadge \\
Strategic behavior of large language models and the role of game structure versus contextual framing \citep{lore_strategic_behavior_2024} & \qibadge \\
Strategic Reasoning with Language Models \citep{gandhi_strategic_reasoning_2023} & \qiiibadge \\
Strategist: Self-improvement of LLM Decision Making via Bi-Level Tree Search \citep{light_strategist_selfimprovement_2025} & \qiiibadge \\
Strategy Adaptation in Large Language Model Werewolf Agents \citep{nakamori_strategy_adaptation_2025} & \qiiibadge \\
Strategy-Augmented Planning for Large Language Models via Opponent Exploitation \citep{xu_strategyaugmented_planning_2025} & \qibadge \\
Stress Testing Deliberative Alignment for Anti-Scheming Training \citep{schoen_stress_testing_2025} & \qiiibadge \\
Super-additive Cooperation in Language Model Agents \citep{tonini_superadditive_cooperation_2025} & \qiiibadge \\
The Surprising Effectiveness of PPO in Cooperative Multi-Agent Games \citep{yu_surprising_effectiveness_2022} & \qibadge \\
Suspicion-Agent: Playing Imperfect Information Games with Theory-of-Mind-Aware GPT-4 \citep{guo_suspicionagent_2024} & \qibadge \\
SwarmBrain: Embodied agent for real-time strategy game StarCraft II via large language models \citep{shao_swarmbrain_embodied_2024} & \qibadge \\
TextArena \citep{Guertler2025textarena} & \qiiibadge \\
Time to Talk: LLM Agents for Asynchronous Group Communication in Mafia Games \citep{eckhaus_time_talk_2025} & \qiiibadge \\
Towards AI-Assisted Generation of Military Training Scenarios \citep{Hans2025assisted} & \qiibadge \\
Towards Strategic Persuasion with Language Models \citep{cheng_towards_strategic_2025} & \qiiibadge \\
TowerMind: A Tower Defence Game Learning Environment and Benchmark for LLM as Agents \citep{Wang2026towermind} & \qibadge \\
Tracing LLM Reasoning Processes with Strategic Games \citep{yuan_tracing_llm_2025} & \qibadge \\
Training Language Models for Social Deduction with Multi-Agent Reinforcement Learning \citep{sarkar_training_language_2025} & \qiiibadge \\
The Traitors: Deception and Trust in Multi-Agent Language Systems \citep{curvo_traitors_deception_2025} & \qiiibadge \\
TRPG Game Mastering Using LLM-Based Multi-Agent System \citep{minari_trpg_game_2025} & \qivbadge \\
UNBench: Benchmarking LLMs for Political Science (UN Security Council) \citep{liang_benchmarking_llms_2026} & \qiiibadge \\
Using Artificial Intelligence Algorithms for High Level Tactical Wargames and New Approaches to Wargame Simulation \citep{lucek_using_artificial_2018} & \qibadge \\
Using Combat Simulations to Determine Tactical Responses to New Technologies on the Battlefield \citep{mittal_using_combat_2024} & \qibadge \\
Verbal Werewolf: Engage Users with Verbalized Agentic Werewolf Game Framework \citep{Fan2025verbal} & \qiiibadge \\
Vox Deorum: A Hybrid LLM Architecture for 4X / Grand Strategy Game AI (Civilization V) \citep{chen_vox_deorum_2025} & \qibadge \\
War and Peace (WarAgent): Large Language Model-based Multi-Agent Simulation of World Wars \citep{hua_war_peace_2024} & \qiiibadge \\
Welfare Diplomacy: Benchmarking Language Model Cooperation \citep{mukobi_welfare_diplomacy_2023} & \qibadge \\
Werewolf Arena: A Case Study in LLM Evaluation via Social Deduction \citep{Bailis2024werewolf} & \qiiibadge \\
WGSR-Bench: Wargame-based Game-theoretic Strategic Reasoning Benchmark for Large Language Models \citep{yin_wgsrbench_wargamebased_2025} & \qiiibadge \\
What if LLMs Have Different World Views: Simulating Alien Civilizations with LLM-based Agents \citep{xue_what_if_2025} & \qivbadge \\
When Two LLMs Debate, Both Think They'll Win \citep{prasad_when_two_2025} & \qivbadge \\
Who Speaks Next? Multi-party AI Discussion Leveraging the Systematics of Turn-Taking \citep{nonomura_who_speaks_2025} & \qiiibadge \\
WOLF: Werewolf-based Observations for LLM Deception and Falsehoods \citep{agarwal_wolf_werewolfbased_2025} & \qiiibadge \\
You Have Thirteen Hours in Which to Solve the Labyrinth: Enhancing AI Game Masters with Function Calling \citep{Song2024have} & \qivbadge \\
ZeroSumEval: Scaling LLM Evaluation with Inter-Model Competition \citep{khan_zerosumeval_scaling_2025} & \qiiibadge \\
\bottomrule
\end{longtable}
\end{center}

\makeatletter
\if@twocolumn\twocolumn\fi
\makeatother

\subsection{Q-IV Anchor: Studies with Creative Players and Creative Adjudicators}
\label{sec:qiv-anchor}

\begin{figure}[t]
\centering
\includegraphics[width=\linewidth]{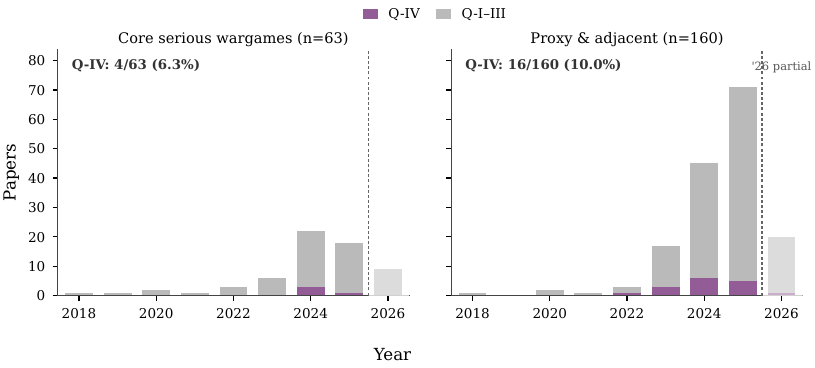}
\caption{Q-IV (both-creative) studies by corpus stratum. Within core serious wargames (left), Q-IV accounts for 4 of 63 studies ($\approx$6.3\%)---a smaller share than among proxy and adjacent studies (right; 16 of 160, 10.0\%) or the corpus overall (20 of \numpapers; $\approx$9\%). The regime serious wargaming most relies on is the one its own literature exercises least. 2026 counts are partial (papers through May 1).}
\label{fig:qiv-case}
\end{figure}

The 20 main-corpus papers placed in Q-IV (HIGH player creativity, HIGH adjudicator creativity) are listed in \reftab{qiv-anchor} with a one-sentence context derived from the released artifact's curation rationale. Three of the 20 (\citet{engels_scaling_laws_2025}, \citet{xue_what_if_2025}, \citet{prasad_when_two_2025}) sit at the boundary of the serious-wargame definition (\refsec{background}); the corresponding strict-versus-permissive recount appears in \refsec{results}. \reffig{qiv-case} situates these studies against the rest of the corpus over time: even as AI-in-wargames work grows, Q-IV stays a thin band. \reftab{core-proxy} stratifies the quadrant distribution by corpus stratum: within core serious wargames the Q-IV share falls to 4 of 63 ($\approx$6.3\%), below both the proxy-study share (16 of 160; 10.0\%) and the full-corpus share ($\approx$9\%).

\begin{table}[t]
\centering
\caption{Quadrant distribution by corpus stratum. Core serious wargames are the 63 main-corpus papers carrying the military-wargame domain flag in the released catalog; proxy and adjacent studies are the remaining 160. Q-IV---the regime serious wargaming most relies on---is scarcer within core serious wargames (4 of 63; $\approx$6.3\%) than in the corpus overall (20 of \numpapers; $\approx$9\%).}
\label{tab:core-proxy}
\begin{tabular}{lrrr}
\toprule
Quadrant & Core & Proxy & Full corpus \\
\midrule
\qibadge   & 42 & 40  & 82 \\
\qiibadge  & 5  & 6   & 11 \\
\qiiibadge & 12 & 98  & 110 \\
\qivbadge  & 4  & 16  & 20 \\
\midrule
Total      & 63 & 160 & \numpapers \\
Q-IV share & 6.3\% & 10.0\% & 9.0\% \\
\bottomrule
\end{tabular}
\end{table}

\begin{center}
\small
\begin{longtable}{p{0.30\linewidth}p{0.62\linewidth}}
\caption{\qivbadge main-corpus papers: studies in which both the player and the adjudicator are coded HIGH on creativity. Context strings are condensed from the curation rationale field of the released bibliography artifact's spreadsheet.}\label{tab:qiv-anchor}\\
\toprule
\textbf{Reference} & \textbf{Context} \\
\midrule
\endfirsthead
\toprule
\textbf{Reference} & \textbf{Context} \\
\midrule
\endhead
\citet{callison-burch_dungeons_dragons_2022} & \gameDnD as an open-ended collaborative role-play and storytelling task with player--DM dialogue and negotiation. \\
\citet{liang_encouraging_divergent_2024} & Structured multi-agent debate framework where LLMs argue opposing positions. \\
\citet{zhou_cast_detect_2023} & \gameDnD DM guidance as a creative-adjudicator/teacher role; player responses and table interaction are natural-language and open-ended. \\
\citet{zhou2023sotopia} & LLM-as-game-player setting for negotiation, social-deduction, and multi-party games. \\
\citet{chen_large_language_2024} & Experiments deploying LLMs in a commercial military wargame with adversarial robustness testing. \\
\citet{weller_leveraging_large_2024} & NATO STO wargaming prototype using LLMs for content generation, facilitation, information provision/action masking, and wargame support. \\
\citet{Kenton2024scalable} & Multi-agent debate/consultancy setting where LLM debaters produce arguments and a weaker LLM judge evaluates them. \\
\citet{hogan_openended_wargames_2024} & IQT agentic architecture for using LLMs in qualitative wargaming. \\
\citet{thudium_outwit_outplay_2024} & Social-competition game similar to \gameSurvivor, with alliance-building, persona-driven play, and elimination voting. \\
\citet{Peng_2024} & LLM-driven text-adventure/NPC framework studying emergent player-driven narrative. \\
\citet{engels_scaling_laws_2025} & Scaling laws for oversight, including via debate-game settings. \\
\citet{xue_what_if_2025} & Open-ended LLM multi-agent simulation of alien civilizations. \\
\citet{Song2024have} & LLM AI Game Master combining narration and function calling to maintain state and adjudicate free-form player actions. \\
\citet{chuang_debate_largescale_2026} & Multi-round role-playing LLM-agent debate/social simulation with public discourse and private opinion-state dynamics. \\
\citet{elbaum_managing_escalation_2025} & Escalation dynamics of LLMs deployed in military/crisis wargame contexts. \\
\citet{zeng_setting_dc_2025} & LLM agents participate in \gameDnD simulations with tool-grounded rules and role-specific play. \\
\citet{jorgensen_static_vs_2025} & Agentic Game Master AI facilitates solo RPG play with interactive text-based narrative; free-form player actions and creative AI-GM adjudication. \\
\citet{minari_trpg_game_2025} & LLM multi-agent system automates game-master tasks for tabletop role-playing games, including scenario, rule, and narrative management. \\
\citet{prasad_when_two_2025} & LLM overconfidence in two-LLM debate self-evaluation. \\
\citet{Anantaprayoon2026learning} & Multi-agent deliberation/negotiation exploring collective value alignment with open-ended arguments and outcome evaluation. \\
\bottomrule
\end{longtable}
\end{center}

\subsection{Domains in Wargames}
\label{sec:wargame_domains}

This section synthesizes key design principles for developing and evaluating LM-driven agents in open-ended wargames. Drawing from our scoping review, we distill a set of core methodological considerations---such as turn structure, evidence requirements, human facilitation, and adjudication protocols---that influence the validity and analytical utility of language-based strategic simulations \citep{downes-martin_validity_utility_2017}. We then contextualize these principles within specific domains to provide actionable guidance for researchers.

\subsubsection{Military and National Security}

AI offers militaries and national security establishments several new training and planning methodologies. Multiple defense organizations are actively exploring how to use AI to provide experiential learning and establish strategic advantage through superior decision-making and judgment \citep{black_scaling_artificial_2024}. Wargames offer humans a simplified mental model to navigate decision-making and analyze results, so they have been used extensively in military education and training \citep{us_naval_war_college_war_gamers_2015, uk_ministry_of_defense_wargaming_handbook_2017, us_army_how_master_2023}.

Both military SMEs and academic researchers are currently investigating how LMs can act as players and adjudicators in open-ended wargames \citep{black_mastering_digital_2024, griffin_matrix_gaming_2024}. We have identified several opportunities to infuse LMs into these activities:
\begin{itemize}
    \item \textbf{Tactical Level:} While LMs are unlikely to take to the battlefield as an opposing force, LM wargaming software could be connected with existing battle command systems to increase staff engagement during field training exercises.
    \item \textbf{Operational Level:} Staff wargaming activities can be greatly enhanced by having an LM agent play the role of the opposing forces commander, adjudicate, or play out unexplored branches and sequels from decision points.
    \item \textbf{Strategic Level:} LMs could explore branches and sequels more broadly than human-only exercises permit; whether that exploration surfaces probable outcomes remains untested in the reviewed corpus.
    \item \textbf{Grand Strategic Level:} LMs could be tasked with exploring plans and policy outcomes; no study in the reviewed corpus evaluates whether doing so confers a decision-making advantage.
\end{itemize}

\subsubsection{Cybersecurity}
While cybersecurity is a distinct domain from military and national security, it is inherently adversarial. Cybersecurity wargames are easily separated into technical, operational, and policy issues \citep{samuelson_wargaming_cybersecurity_2018}, and LMs provide distinct value in each:
\begin{itemize}
    \item \textbf{Technical Wargames:} These test the skills of hands-on operators. LMs provide value as AI teammates or adversaries, or as adjudicators for inputs beyond on-network computer commands.
    \item \textbf{Operational Wargames:} These serve both technical teams and non-technical organizations. For technical teams, LMs can extend the on-network narrative to security analysts by generating digital content that represents or reacts to on-net activity. For non-technical organizations evaluating incident response plans, LMs can generate cohesive content across formats (system logs, blog posts, government advisories) and process disparate player activities for adjudication.
    \item \textbf{Policy Wargames:} Policymakers do not need to understand technical details, but the wargame must remain technically accurate. LMs can maintain an accurate representation of the technical state while processing plain language back and forth between players and the gaming platform.
\end{itemize}

\subsubsection{International Relations}
International Relations (IR) wargaming centers on language, signaling, and credibility rather than material force. Formats like seminar and matrix games capture how positions evolve through persuasion, norm invocation, and threat--promise exchanges \citep{mans_training_warriordiplomat_2010, schechter_wargaming_methodology_2021, lin-greenberg_wargaming_international_2022}. Wargames are commonplace among IR professionals to train and analyze unseen diplomatic crises (e.g., nuclear crises) \citep{hersman_nuclear_shadow_2020, reddie_evidence_unthinkable_2023, worman_designing_strange_2023}. 

When LMs participate in diplomatic games, they can assist with the unique challenges of incomplete information, face-saving, and issue linkage by:
\begin{itemize}
    \item \textbf{Narrative Maintenance:} Maintaining coherent narratives over long horizons and proposing plausible options under contested facts.
    \item \textbf{Content Generation:} Drafting position papers, back-channel messages, or press releases in different registers, while humans retain control of red lines and legitimacy constraints.
    \item \textbf{Adjudication Support:} Helping surface implicit assumptions in briefs and aiding with adjudication on subjective, non-analytical rubrics.
\end{itemize}

\subsubsection{Social Games}

While wargames may contain elements of social reasoning or deception, they differ fundamentally from social deduction games. Wargames require agents to pursue defined goals through planning within a structured conflict environment, whereas social deduction games center on identity discovery and psychological misdirection.

Psychological misdirection, however, is prolific in warfare (e.g., Trojan Horse, Empty Fort Strategy). For this review, social deduction games represent a complementary effort to wargames:
\begin{itemize}
    \item \textbf{Navigating Deception:} Social deduction games test LMs' ability to navigate dynamic, multi-agent interactions involving trust and deception \citep{chi_amongagents_evaluating_2024}.
    \item \textbf{Real-World Transfer:} Harnessing psychological misdirection in game scenarios enhances LM performance in real-life contexts where information may be unreliable or intentionally deceptive \citep{maggio_game_theory_2024, lamparth_human_vs_2024}.
\end{itemize}

\subsubsection{Economics and Business}

In business contexts, wargaming draws on competitive strategy to stress-test hypotheses during adversarial dynamics, market shifts, and policy shocks \citep{hamel_competing_future_1994, bradfield_origins_evolution_2005}. Seminar-style sessions combine narrative role-play (competitor, regulator) with structured turns (pricing, product launch) \citep{kurtz_business_wargaming_2003}. LM assistance supports scale and reproducibility in these games by:
\begin{itemize}
    \item \textbf{Ideation:} Providing speed and breadth in enumerating strategic options and probing counterfactuals (e.g., ``what would a rational competitor do if...'') \citep{chussil_learning_faster_2007}.
    \item \textbf{Roleplay and Content:} Drafting memos in given styles and modeling specific economic actor behaviors.
    \item \textbf{System Modeling:} Deploying AI agents as firms or traders allows the study of collusion, innovation, shocks, and equilibria under controlled conditions.
\end{itemize}

Wargames also model economic systems directly (e.g., \gameCivilizationIV, \gameCatan). Decisions operationalize core concepts like opportunity cost, inflation, deficit spending, and transaction costs \citep{alabdulkarim_goaldirected_story_2021}. Evaluating reinforcement-learning and LM agents in such settings provides an experimental platform for testing assumptions and measuring second-order effects of policy initiatives.

\subsubsection{Medicine and Public Health}

Wargaming has also been used to great effect in medicine and public health to prepare for and respond to health crises. 
\begin{itemize}
    \item \textbf{Preparedness Testing:} Exercises like \gameDarkWinter (simulating a smallpox attack) and \gameEventTwoZeroOne (simulating a global pandemic) are tabletop exercises (TTXs) that test emergency plans and improve inter-agency coordination.
    \item \textbf{Rehearsing Responses:} These exercises are indispensable in preparing for real-life events, allowing policymakers and healthcare professionals to rehearse responses in a controlled environment \citep{smith_serious_games_2020}.
\end{itemize}

\section{Detailed Recommendations}\label{sec:detailedrecommendations}

\textbf{Task-specific baselines comparison.} Establishing control conditions using deterministic agents or human SME players or adjudicators enables qualitative and quantitative measurements of LM agent performance in various conditions, and can help detect systematic biases or failure modes unique to LM reasoning \citep{yin_wgsrbench_wargamebased_2025}. Existing human baselines in relevant task spaces (e.g., creative writing, strategic deception) are largely neither transparent nor rigorous enough to provide meaningful comparisons \citep{wei_position_human_2025}. High-stakes wargames, therefore, should prioritize bespoke evaluations with scenario-relevant metrics and adequate analysis to identify capability gaps and boundary conditions before operational use \citep{lin-greenberg_wargame_drones_2022, caballero_large_language_2025, chu_domaino1s_guiding_2025, tang_dsgbench_diverse_2025}.

\textbf{Robustness testing.} To measure LM reliability, running inference across paraphrased inputs, synonym substitutes, and varied prompt structures may surface inconsistent strategic reasoning \citep{shrivastava_measuring_freeform_2024, nalbandyan_score_systematic_2025}. Testing both surface-level syntactic robustness and semantic equivalence can largely be automated using auxiliary and smaller LMs and integrated into deployed workflows to inform user confidence in outputs.

\textbf{Calibration assessment.} Models with well-calibrated confidence help avoid overreliance on flawed strategic assessments and under-reliance on sound reasoning, providing an important auditing mechanism for understanding LM decisions; measurements of LM calibration allow external stakeholders of wargames to understand systematic flaws in LM decision-making. Additionally, requiring LMs to quantify uncertainty is likely to improve agent performance and make human review of key actions more efficient, particularly in high-stakes situations \citep{liu_uncertainty_quantification_2025, downes-martin_preference_reversal_2020, freeman_artificial_intelligence_2024}.

\textbf{Validation robustness.} LMs reliably detect evaluation contexts and may perform differently when aware they are being tested \citep{needham_large_language_2025, abdelnabi_linear_control_2025}, potentially masking real-world failure modes or displaying deceptive reasoning during assessment. Multiple model architectures should be tested on identical scenarios to identify high-uncertainty areas and common failure modes, while evaluation awareness should be monitored through motivated questioning (``Do you believe you are being evaluated?'') and passive CoT analysis to improve performance. For instance, cross-model critique, while underperforming external feedback \citep{gou_critic_large_2024}, outperforms self-correction in multi-agent settings \citep{saleh_evaluating_large_2025}. Episodes with evaluation awareness should be reevaluated, and significant consensus breakdowns may signal events requiring human oversight.

\textbf{Human stakeholder training.} LMs' non-intuitive failure modes do not align with the expectation of stakeholders, who are likely to ascribe moral intent to LM output and unlikely to question plain statements \citep{sharma_why_would_2024}. Operators need a technical understanding of when to trust, how to improve, and where to audit LM outputs. Key stakeholders, including decision-makers who rely on LM-enabled wargames, should understand LM behavioral markers and be provided with confidence assessments of wargame conclusions \citep{ehsan_humancentered_explainable_2020}.

\section{Extended Scope and Released Artifact}
\label{sec:artifacts-llm}

\subsection{Extended Scope Notes}
\label{sec:extended-scope}

The body limitations section (\refsec{limitations}) covers coding stability. Three further scope edges---boundary cases on the continuous axis, candidate future axes, and the corpus's coverage window---are recorded here for readers who would extend the rubric.

\paragraph{Boundary cases on a continuous axis.}
The HIGH/LOW split simplifies a continuous space. The boundary is sharp for canonical cases---\gameNoPressDiplomacy is unambiguously LOW on player creativity; tabletop role-playing exercises are unambiguously HIGH on adjudicator creativity---but it blurs in open-world settings whose interaction surface is language-rich while the action set narrows toward a fixed terminal goal. As a worked example, an open-world role-playing game such as \gamePokemon affords extensive natural-language interaction with NPCs but constrains the player to a small set of structurally similar actions and a single terminal objective; we code such studies LOW on player creativity to remain consistent with the rest of the corpus, but a different coder applying the same definitions could reasonably disagree. Future work should publish a richer worked-example boundary set than the anchors offered by the present rubric.

\paragraph{Game protocol versus AI deployment.}
Our rubric describes properties of the wargame itself---what its protocol permits players to say and what it permits adjudicators to decide---rather than how AI is configured inside that protocol. Two studies in the same quadrant can therefore differ along at least two further dimensions. The first is \emph{AI directionality}: which side of the player--adjudicator dyad hosts the AI (AI-vs-AI, AI-vs-human, or peer-ranked LLM-as-judge variants). The second is \emph{AI agency level}: for each AI-bearing role, whether the AI is an assistant to a human or the autonomous agent acting in that role. A human SME validating LLM-generated adjudications is not the same setup as an LLM acting as the adjudicator, even though both can be coded HIGH on adjudicator creativity.

These dimensions cut along the broader Human-AI (HAI) versus AI-vs-AI (A2A) distinction. A2A is increasingly important as agentic AI is deployed in organizational settings, but competitive A2A diplomacy is underexplored, and human-likeness---a standard HAI benchmark---may not transfer cleanly to A2A settings with distinct equilibria. We did not code these dimensions for the present corpus, and they are not currently recorded as per-paper flags in the released artifact; we flag them as candidate axes for a future rubric extension.

\paragraph{Corpus scope.}
The review is limited to unclassified work with available full text and therefore cannot capture classified wargaming practice, internal evaluations, or unpublished operational lessons. The full main corpus contains \numpapers~de-duplicated studies retrieved through May 1, 2026.

\subsection{Released Artifact}
\label{sec:released-artifact}

We include the reviewed-paper database as an anonymized supplementary artifact. The artifact contains the full included bibliography, retained near-miss audit records, metadata, search dates, databases searched, de-duplication notes, post-hoc analytic flags, coding fields, curation rationales, and notes used to support the scoping review and rubric application. It also includes documentation describing the metadata fields, inclusion criteria, exclusion codes, coding rubric, manual curation steps, intended uses, limitations, and review license terms. If accepted, we will release a de-anonymized version under a permissive public license at a permanent archive.

\end{document}